\documentclass[final,3p,times]{elsarticle}

\usepackage{amssymb}
\usepackage{graphicx,multirow,caption,subcaption,multicol,setspace,amsmath,wrapfig,makecell}
\usepackage[linesnumbered,ruled]{algorithm2e}
\usepackage{algorithmic}
\usepackage{adjustbox}
\usepackage{booktabs,caption}
\usepackage[flushleft]{threeparttable}
\usepackage{lineno}
\usepackage{enumitem}
\setlist{nosep}
\usepackage{rotating,caption}
\usepackage[table]{xcolor}
\usepackage{soul,xcolor}
\usepackage{amsfonts}
\usepackage{pifont}
\newcommand{\xmark}{\ding{55}}%
\usepackage{amsthm}
\theoremstyle{definition}

\biboptions{numbers,sort&compress}
\graphicspath{ {Images/} }

\makeatletter
\newcommand{\algorithmfootnote}[2][\footnotesize]{%
  \let\old@algocf@finish\@algocf@finish% Store algorithm finish macro
  \def\@algocf@finish{\old@algocf@finish% Update finish macro to insert "footnote"
    \leavevmode\rlap{\begin{minipage}{\linewidth}
    #1#2
    \end{minipage}}%
  }%
}

\newcommand{\etal}{\textit{et al.}}

\journal{Applied Soft Computing}

\begin{document}

% \linenumbers

\begin{frontmatter}

\title{Efficient Feature Selection of Power Quality Events using Two Dimensional (2D) Particle Swarms}

\author[label5]{Faizal Hafiz\corref{cor1}}
\address[label5]{Department of Electrical \& Computer Engineering, The University of Auckland, Auckland, New Zealand}
\ead{faizalhafiz@ieee.org}
\cortext[cor1]{Corresponding author}

\author[label5]{Akshya Swain}

\author[label1]{Chirag Naik}
\address[label1]{Sarvajanik College of Engineering \& Technology, Surat, India}

\author[label5]{Nitish Patel}

\begin{abstract}

A novel two-dimensional (2D) learning framework has been proposed to address the feature selection problem in Power Quality (PQ) events. Unlike the existing feature selection approaches, the proposed 2D learning explicitly incorporates the information about the subset cardinality (\textit{i.e., the number of features}) as an additional learning dimension to effectively guide the search process. The efficacy of this approach has been demonstrated considering fourteen distinct classes of PQ events which conform to the IEEE Standard 1159. The search performance of the 2D learning approach has been compared to the other six well-known feature selection wrappers by considering two induction algorithms: \textit{Naive Bayes} (NB) and \textit{k-Nearest Neighbors} (k-NN). Further, the robustness of the selected/reduced feature subsets has been investigated considering seven different levels of noise. The results of this investigation convincingly demonstrate that the proposed 2D learning can identify significantly better and robust feature subsets for PQ events.

\end{abstract}


\end{frontmatter}

\section{Introduction}
\label{sec1:intro}

Over the past two decades, the landscape of the energy market has been going through a significant transformation due to the increasing share of non-linear loads. The breakthrough progress in the semiconductor technology has enabled a wide-scale deployment of the power electronic converters, adjustable speed drives and consumer electronics. Furthermore, the power electronic converters are the key components in the grid interface of the renewable generation. In this scenario, one of the major challenge faced by the utilities is the deterioration in Power Quality (PQ). Since the success of the remedial action is critically dependent on the nature of the PQ event, the identification of a PQ event is vital in addressing the poor PQ. This study, therefore, focuses on identification of PQ events.

In essence, any departure from the ideal \textit{sinusoid} can be considered as a PQ event. Various PQ events exist in the modern grid which differ in terms of \textit{spectral content, magnitude} and \textit{duration}. Based on these \textit{`traits'}, the IEEE standard 1159 characterizes PQ events of distinct nature~\cite{IEEE:1159}. Arguably, the earliest attempt to identify the PQ events could be traced back to the works of Santoso \textit{et al.}~\cite{Santoso:1996,Santoso:2000}, Angrisani \textit{et al.}~\cite{Angrisani:1998} and Gaouda \textit{et al.}~\cite{Gaouda:Salama:1999}, in which the feature extraction through multi-resolution analysis was proposed. Subsequently, over the years, the PQ event identification problem has transformed into a \textit{pattern recognition} problem wherein a \textit{pattern} is formed by various \textit{features} extracted from the voltage measurements and a \textit{label} corresponding to PQ event, \textit{i.e.}, $i^{th}$ pattern, `$\ell_i$', is given by, $\ell_i=\{ u_1^i, u_2^i, \dots u_n^i, v_i\}$, where, `$v_i$' and $\{ u_1^i, u_2^i, \dots u_n^i \}$ respectively represent the \textit{output label} and the features extracted from the $i^{th}$ PQ event. The patterns extracted from the PQ events are subsequently used to induce a \textit{classifier} through a suitable induction algorithm. Most of the existing PQ event identification approaches are based on this framework~\cite{Mahela:Shaik:2015,Khokhar:Zin:2015}. Note that the utility/efficacy of the extracted features is usually not evaluated in most of the existing PQ identification approaches~\cite{Mahela:Shaik:2017,Kumar:Singh:2015,Liu:Cui:2015,Biswal:Dash:2013,Naik:Hafiz:2016}, which often leads to the inclusion of irrelevant, redundant and noisy features.

Despite the significant research efforts dedicated to the machine learning and pattern recognition, the feature extraction process is still, in major part, dependent on the expert knowledge. Since it is not trivial to estimate the required number of features/attributes beforehand, a large number of features are usually extracted in order to identify a \textit{pattern}. This often leads to the inclusion of irrelevant and/or redundant features which often lead to increased storage requirement, slower processing times and reduced generalization capability of the classifier. The objective of the \textit{feature selection} is to overcome these problems by the removal of \textit{irrelevant/redundant} features. Note that the feature selection is one of the fundamental problems of the machine learning and it has been topic of active research since last six decades~\cite{Marill:Green:1963,Blum:Langley:1997,Dash:Liu:1997,Guyon:Isabelle:2003,Xue:Zhang:2016}. Over the years, several feature selection approaches have been developed with a proven ability to significantly reduce the number of features while maintaining/improving classification performance in many applications~\cite{Guyon:Isabelle:2003,Xue:Zhang:2016}. 

Despite these advantages, the feature selection has not received enough attention in PQ identification except for a few existing research~\cite{Panigrahi:2009,Gunal:2009,Lee:Shen:2011,Manimala:Selvi:2011,Manimala:Selvi:2012,Ericsti:2013,Dalai:Chatterjee:2013,Hajian:Foroud:2014a,Hajian:Foroud:2014b,Abdoos:Mianaei:2016,Hafiz:Swain:2017,Khokhar:Zin:2017,Singh:Singh:2017}. The choice of the feature selection method varies among the existing research which include \textit{Sequential Search}~\cite{Gunal:2009,Hajian:Foroud:2014a,Hajian:Foroud:2014b,Abdoos:Mianaei:2016}, \textit{Genetic Algorithm} (GA) \cite{Panigrahi:2009,Gunal:2009,Manimala:Selvi:2011,Manimala:Selvi:2012,Hafiz:Swain:2017}, \textit{Simulated Annealing} (SA)~\cite{Manimala:Selvi:2011,Manimala:Selvi:2012}, \textit{Binary Particle Swarm Optimization} (BPSO)~\cite{Hafiz:Swain:2017}, \textit{Fully Informed Particle Swarm} (FIPS) \cite{Lee:Shen:2011}, \textit{Artificial Bee Colony} (ABC) \cite{Khokhar:Zin:2017}, \textit{k-means apriori algorithm} \cite{Ericsti:2013} and \textit{rough sets} \cite{Dalai:Chatterjee:2013}. The common drawback of the sequential search method such as \textit{Sequential Forward Search} (SFS) and \textit{Sequential Backward Search} (SBS) is the so-called `\textit{nesting effect}', \textit{i.e.}, once the feature is included/excluded from the subset, it cannot be removed/added. The floating sequential search approaches such as \textit{Plus-l Take Away-r} (PTA) and \textit{Sequential Forward Floating Search} (SFFS) can prevent nesting effect. However, these are still \textit{local} search, as the over-emphasis on a single feature neglects the correlation among features. Furthermore, all sequential search methods require specification of the reduced subset size \textit{a priori}. Given that the size of the optimal feature subset is not known, the sequential search essentially addresses only half of the feature selection problem. The meta-heuristic search methods such as GA, SA and ABC could overcome these limitations. However, in~\cite{Panigrahi:2009,Manimala:Selvi:2012,Khokhar:Zin:2017}, these approaches have not been used to full potential; as the size of the feature subset is fixed \textit{a priori}. Such practice is usually not recommended, since fixing the subset size to a particular value, \textit{say} `$d$', will reduce feature search space from $2^n$ to $ \binom {n}{d}$ which may exclude the optimal feature subset from the search. The \textit{filter} based feature selection approaches have been employed in~\cite{Hajian:Foroud:2014a,Hajian:Foroud:2014b,Dalai:Chatterjee:2013} which usually involves a trade-off in the classification performance as the nature of the induction algorithm is not considered. This is further, corroborated by the results of the earlier investigation~\cite{Hafiz:Swain:2017} which suggest that the indirect performance indicators used in the filter approach do not necessarily translate into equivalent classification performance. In~\cite{Lee:Shen:2011}, Probabilistic neural network-based Feature Selection (PFS) has been proposed which employs a combination of Fully Informed Particle Swarm and Adaptive Probabilistic Neural Network to evaluate the efficacy of a feature subset. The major drawback of this approach is the assumption of the \textit{monotonic} relation between the feature subset size and classification performance. This assumption can only be satisfied by an ideal Bayes classifier. In practice, the classifiers do not follow the monotonicity property~\cite{Siedlecki:Sklansky:1989,Kohavi:1994,Kohavi:John:1997,Yang:Honavar:1998}. In addition, PFS is prone to the nesting effect, as it does not have any mechanism to reconsider the discarded features.

Further, in most of the existing research, the \textit{deterministic} feature selection approaches have been included in the comparative analysis whereas it is known that meta-heuristic search could yield enhanced search performance~\cite{Xue:Zhang:2016}. However, among the meta-heuristic search, so far only the performance of GA, SA and BPSO have been evaluated~\cite{Panigrahi:2009,Manimala:Selvi:2011,Manimala:Selvi:2012,Hafiz:Swain:2017}. In this scenario, the following questions arise:
% \smallskip
\begin{enumerate}
    \smallskip
    \item \textit{Amongst various meta-heuristic approaches (such as GA, BPSO, ACO and others) which approach would yield the minimum feature subset without compromising the classification performance?}
    \smallskip
    \item \textit{Whether the classification performance obtained from the reduced feature subset is robust against various levels of measurement noise?}
    \smallskip
\end{enumerate}
\smallskip 

The main aim of this study is to address these concerns through a comprehensive investigation. For this purpose, a comparative analysis of seven different feature selection wrappers is carried out using fourteen distinct class of PQ events and two different induction algorithms. Further, the robustness of the reduced subsets against the measurement noise is evaluated through seven different levels of \textit{zero-mean Gaussian white noise}.

The other objective of this study is to investigate the efficacy of the new Two-Dimensional feature selection algorithm on the PQ events. The Two-Dimensional (2D) learning algorithm has recently been developed by the authors as a generalized feature selection wrapper~\cite{Hafiz:Swain:2017a}. The core idea of the 2D-learning approach is to embed the information about \emph{the number of features} (referred to as \emph{`cardinality'}) into the learning process of particle swarms. This distinctive quality of the 2D learning approach has been shown to be more effective in identifying the compact feature subsets with the improved classification performance for several benchmark machine learning datasets. This, therefore, has been the motivation behind the application of 2D learning to the PQ event identification problem. In particular, the present study focuses on the practical issues encountered in the feature selection of PQ events and therefore significantly differs from~\citep{Hafiz:Swain:2017a} in the following aspects: 
\medskip
\begin{itemize}
    \item A well-balanced, comprehensive and diverse PQ event dataset is built. The dataset includes both simulated and experimental PQ events which conform to the IEEE Std. 1159. Further, a relatively large variation of PQ events is considered in comparison to the existing research on PQ identification~\cite{Lee:Shen:2011,Biswal:Dash:2013,Biswal:Dash:2013a,Kumar:Singh:2015,Singh:Singh:2017,Mahela:Shaik:2017}. For example, the PQ events include magnitude variation in the range of $[0, \ 4] \ pu$, event duration from $0.5$ to $30$ cycles, frequency variation in the range of $[0, \ 10] \ kHz$, and the harmonics of the order $(6k \pm 1), \ k=\{ 1, 2, 3, \dots \}$. Thus, most of the PQ events expected in the conventional power distribution grid are well represented in the investigated dataset.
    \smallskip
    \item The robustness of the selected/reduced feature subsets against the measurement noise is defined and evaluated under various levels of zero-mean Gaussian white noise.%: $[50dB, 45dB, 40dB, 35dB, 30dB, 25dB, 20dB]$ SNR.
    \smallskip
    \item The Two-Dimensional (2D) learning algorithm is better formalized and explained through an illustrative example from the perspective of PQ event identification.
    \smallskip
    \item The efficacy of 2D learning algorithm is demonstrated by the comparative evaluation on six established feature selection algorithms: GA, ACO, BPSO, CBPSO, ChBPSO and SFFS.
    \smallskip
    \item A rigorous analysis is carried out to determine the statistical significance of the results. For instance, multiple non-parametric statistical comparisons are carried out using the Friedman test and the Hommel's post-hoc procedure to compare the search performance of the algorithms. In addition, the performance of algorithms under various levels of noise is compared using the Contrast Estimation based on medians. 
\end{itemize}

Thus, in this study, most of the operating scenarios expected in the utility distribution grid have been accommodated. Further, several key issues related to the feature selection have been investigated.

The rest of the article is organized as follows: Section~\ref{s:fm} provides a brief overview of the feature selection approaches. The 2D-learning approach has been briefly discussed in Section~\ref{s:2DL}. The investigation framework of this study is provided in Section~\ref{s:IF}. The results of the comparative evaluation and the robustness test are shown in Section~\ref{s:res}, followed by the discussions in Section~\ref{s:discuss} and the conclusions in Section~\ref{s:con}.

%-------------------------- New Section ---------------------------------
\section{Brief Overview of Feature Selection Methods}
\label{s:fm}
Since the main objective of this work is to select the relevant feature subset for the PQ event identification, it is pertinent to discuss the existing feature selection methods briefly.

The feature selection problem essentially begins by considering a dataset having `$n$' input features, $U=\{u_1,\dots u_n\}$ and `$m$' output class, $V=\{v_1,\dots v_m\}$. For this dataset, the task of the induced \textit{classifier} is to determine the output label `$v_k$' ($v_k\in V$) corresponding to the input features. The objective of the feature selection is to identify the subset of features, `$\mathcal{X}^{\star}$', ($\mathcal{X}^{\star}\subset U, \; \xi_{\mathcal{X}^{\star}}<n$) through which this task can be accomplished with similar or improved classification performance:
%---------------------------------------------------------
\begin{linenomath*}
\begin{align} 
% \label{eq:1}
    J(\mathcal{X}^{\star}) = \max \limits_{\mathcal{X} \subset U, \ \xi_\mathcal{X}<n} J(\mathcal{X}) 
\end{align}
\end{linenomath*}
%--------------------------------------------------------- 
where, `$J(\cdotp)$' is the criterion function which represents the classification performance of the feature subset and `$\xi$' denotes the \textit{cardinality} or \textit{number of features} in the subset. The search for an optimal solution of the feature selection requires the evaluation of $2^n$ subsets. Hence the exhaustive search is intractable even for the moderate size datasets \cite{Cover:Van:1977}. Note that the selection of a relevant feature subset from the available features is essential for the better classification performance and the generalization capability of the classifier \cite{Blum:Langley:1997}.

Over past six decades, while the fundamental objective of the feature selection problem has not changed (\textit{i.e.}, \textit{identify a subset of relevant features}), the machine learning algorithms have evolved significantly with time to meet the requirements of different applications. For example, in many applications, the class `\textit{labels}' may not be available completely or partially~\cite{Guyon:Isabelle:2003,Liu:Yu:2005}. Such scenarios require unsupervised/semi-supervised machine learning approach. A distinct feature selection strategy is required for each machine learning approach such as supervised or unsupervised learning~\cite{Guyon:Isabelle:2003,Liu:Yu:2005,Shang:Wang:2016,Shang:Wang:2018}. In this study, we focus on the feature selection approaches for supervised learning; as the nature of PQ events have been well characterized by the IEEE Std. 1159~\cite{IEEE:1159}. Hence, a comprehensive PQ event dataset has been developed through a combination of simulation and practical field measurements for the supervised learning.

Most of the existing feature selection methods for supervised learning can be distinguished by their approach (\textit{e.g.}, \textit{wrapper} or \textit{filter}) to evaluate the criterion function, $J(\cdotp)$ \cite{Blum:Langley:1997,Dash:Liu:1997,Kohavi:John:1997}. The \textit{wrapper} approach is straightforward where a classifier is induced for each feature subset under consideration and the resulting classification accuracy is used as $J(\cdotp)$. On the contrary, in the \textit{filter} methods, $J(\cdotp)$ is estimated using statistical or information theory without inducing a classifier. Note that the search landscape of the feature selection problem is conjointly defined both by the dataset and the induction algorithm, \textit{i.e.}, each induction algorithm has specific traits. Hence, the optimal feature subset for particular induction algorithm may not be optimal for the other, as will be shown by the results of this study (Section~\ref{s:ce}). 

The choice of feature selection method involves a trade-off in either speed or accuracy; the \textit{wrappers} are more precise whereas \textit{filters} are comparatively faster. The size of the feature set plays a major role in the selection of either approach. For the smaller to medium size feature set ($n<1000$) \textit{wrappers} are more appropriate. For the larger dataset, the computational burden of the wrapper may be infeasible. Hence filters are preferred in such a scenario. Usually, the PQ event datasets range in small to medium size ($n<500$), thus in this work, the focus is on the wrapper approach.
 
Earlier approaches of feature selection, such as \textit{sequential search} methods \cite{Marill:Green:1963,Whitney:1971,Pudil:1994,Somol:Pudil:1999} and \textit{branch and bound} methods \cite{Narendra:1977,Yu:Yuan:1993} are `\textit{deterministic}' in nature, \textit{i.e.}, for a given dataset, they give the same solution over independent runs. The core idea behind most of these approaches is to evaluate the utility/relevance of a feature by evaluating its discrimination capability over the output classes. However, due to over-emphasis on an individual feature, the \textit{correlation} among features is neglected. In addition, most of the deterministic approaches require \textit{a priori} selection of subset cardinality, `$d$'. Consequently, the search space reduces from $2^n$ to $\binom n d$. Further, several deterministic approaches \cite{Narendra:1977,Yu:Yuan:1993} assume a \textit{monotonic} criterion function, $J(\cdotp)$, which is often impractical.

Meta-heuristic search methods such as Genetic Algorithm (GA), Ant Colony Optimization (ACO), Tabu Search (TS) and Particle Swarm Optimization (PSO) have been applied to the feature selection problem to address the shortcomings of the deterministic approaches~\cite{Xue:Zhang:2016}. Unlike deterministic methods, most of the meta-heuristic search methods operate on feature subsets. Hence, the effects of \textit{feature correlation} are accounted in the search process. Further, the meta-heuristic search is, in essence, a \textit{population}-based search with implicit parallelism which allows comparatively better sampling of the search space. For this reason, we focus on the \textit{meta-heuristic wrappers} in this study.

\section{Two-Dimensional (2D) Learning Framework for Particle Swarms}
\label{s:2DL}

The core idea behind the Two-Dimensional (2D) learning approach has been briefly discussed in the following subsections for the sake of completeness. The detailed discussion about this approach can be found in~\cite{Hafiz:Swain:2017a}. Note that, 2D learning is intended to be a \textit{generalized learning} algorithm for particle swarm based feature selection methods. It is, therefore, possible to adapt most of the existing PSO variants~\cite{Hafiz:Abdennour:2013} in the continuous domain ($x \in \mathbb{R}$) for the feature selection problems ($x \in \mathbb{N}$) following the 2D-learning. However, the results of the study in \citep{Hafiz:Abdennour:2016} indicate that among popular PSO variants, adapted Unified Particle Swarm Optimization (UPSO) \citep{UPSO1} performs comparatively better for the problems in the discrete domain. For this reason, in this work, UPSO has been adapted following the 2D-learning approach and referred throughout the manuscript as `2D-UPSO'. 

%------------------------------------------------------------------------
%------                     Pseudo Code:Learning Set
%------------------------------------------------------------------------
\begin{algorithm}[!t]
    \small
    \SetKwInOut{Input}{Input}
    \SetKwInOut{Output}{Output}
    \SetKwComment{Comment}{*/ \ \ \ }{}
    \Input{Particle Position ($\beta_i$) and Learning Exemplar ($\alpha$)}
    \Output{Learning Sets: $\mathcal{L}_\alpha$ and $\mathcal{L}_i$}
    \algorithmfootnote{`$\wedge$' denotes bit-wise logical `AND' operation. `$\overline{\beta_i}$' denotes logical complement of `$\beta_i$'}
    \BlankLine
    \Comment*[h] {Learning for Subset Cardinality}\\
    Set the cardinality learning sets to an $n$-dimensional null vector, \textit{i.e.},
    $\varphi_{\alpha} = \varphi_{i} = \{ 0, \ 0, \dots, 0 \}$\\
    % \BlankLine
    Determine the cardinality of the learning exemplar ($\alpha$) and the particle position ($\beta_i$):\newline
    $\xi_{\alpha} = \sum \limits_{m=1}^{n} \alpha_{m}$ and $\xi_{i} = \sum \limits_{m=1}^{n} \beta_{i,m}, \ \ m=1,2,\dots n $\\
    \BlankLine
    Set the `$\xi^{th}$' bit of the cardinality learning set, `$\varphi$', to `$1$', \textit{i.e.}, \newline $\varphi_{\alpha,\xi_{\alpha}} = 1$ and $\varphi_{i,\xi_{i}} = 1$
    \BlankLine
    \BlankLine
    \Comment*[h] {Learning for Features}\\  
    \BlankLine
    Evaluate Feature Learning Sets: $\psi_{\alpha} = \{ \alpha \wedge \overline{\beta_i} \}$ and $\psi_{i}=\beta_i$ \\
    Evaluate the final learning sets: $\mathcal{L}_{\alpha} = \begin{bmatrix} \varphi_{\alpha} \\ \psi_{\alpha} \end{bmatrix}$ and $\mathcal{L}_{i} = \begin{bmatrix} \varphi_{i} \\ \psi_{i} \end{bmatrix}$
    \BlankLine
\caption{Evaluation of the learning sets}
\label{alg:learningset}
\end{algorithm}
%---------------------------------------------------------
%---------------------------------------------------------
\subsection{Philosophy of 2D Learning}

The search for the optimal feature subset essentially entails the following two decisions: 1) \emph{How many features should be included?} and 2) \emph{Which features should be included?} The philosophy of the Two-Dimensional (2D) learning is to explicitly embed the information about \textit{subset size} (also referred as \textit{cardinality}) into the search process to effectively address these issues. As the name suggests, the 2D-learning approach extends the \textit{learning dimension} of a particle swarm to integrate the cardinality information into the search process. Since the cardinality of the subset is selected through an informed decision, only the features with higher selection likelihoods are selected and the redundant features are effectively discarded.

To further understand the philosophy of 2D learning, consider a feature selection problem associated with a dataset having `$n$' \textit{number of features}. For this dataset, the position of an $i^{th}$ particle, `$\beta_i$', is represented as an $n$-dimensional binary string as follows:
%---------------------------------------------------------
\begin{linenomath*}
\begin{align}
\label{eq:binaryrepresentation}
    \beta_i & = \begin{bmatrix} \beta_{i,1} & \beta_{i,2} & \dots & \beta_{i,n} \end{bmatrix}, \ \ \beta_{i,m} \in [0,1], \ \ m=1,2,\dots n
\end{align}
\end{linenomath*}
%---------------------------------------------------------
where, the bits with `$1$' indicates that the corresponding feature is selected and the bits with `$0$' indicates otherwise, \textit{i.e.}, if $\beta_{i,m}=1$ then $m^{th}$ feature is selected.

Note that in PSO, the `\textit{learning}' of a particle is accumulated in its velocity. For this reason, in 2D learning, the dimension of particle velocity is extended, and it is represented by a two-dimensional matrix. The objective here is to store the \textit{selection likelihood} of \textit{feature} and \textit{cardinality} (\textit{number of features}) in a distinct dimension. For the problem considered here, the velocity of the $i^{th}$ particle, `$\mathcal{V}_i$', is represented by a two-dimensional matrix of size ($2 \times n$) which is given by,
%---------------------------------------------------------
\begin{equation}
\label{eq:vel}
    \mathcal{V}_i=\begin{bmatrix} p_{11}^i & \dots & p_{1n}^i \\ 
                        p_{21}^i & \dots & p_{2n}^i \end{bmatrix}
\end{equation}
%---------------------------------------------------------
The first row of the velocity matrix stores the \textit{selection likelihood} of the \textit{cardinality} (\textit{i.e.}, \textit{subset size} or \textit{number of features}). For instance, $p_{13}^i=2.45$ implies that the likelihood of including a total of $3$ features in the new position of the $i^{th}$ particle is $2.45$. In contrast, the elements in the second row of $\mathcal{V}_i$ store the \textit{selection likelihood} of the corresponding \textit{features}. For example, $p_{2,3}^i=0.75$ indicates that likelihood of including the $3^d$ feature in the new position of the $i^{th}$ particle is $0.75$.

Note that in order to update the selection likelihoods of cardinality and features it is crucial to extract the beneficial information from the \textit{`learning exemplars'} such as `personal best' ($pbest$) and `neighborhood best' ($nbest$). This is accomplished through `\textit{learning}' process in which a dedicated \textit{`learning set'} ($\mathcal{L}$) is derived from each exemplar. The learning set is also a $(2 \times n)$ matrix, in which the first row corresponds to the \textit{cardinality learning} and the second row corresponds to the \textit{feature learning}. In essence, the learning process extracts the following information from the exemplar and encodes it into a two-dimensional binary learning set `$\mathcal{L}$': 1) \textit{number of features in the exemplar} 2) \textit{features that have been included in the exemplar but not in the particle}.

This is achieved as follows: Let `$\alpha$' denote the learning exemplar of the $i^{th}$ particle, \textit{e.g.}, `$pbest_i$', `$nbest_i$' or `$gbest$'. Note that the learning exemplar,`$\alpha$', is also an $n$-dimensional binary string similar to particle position and essentially encodes a feature subset. Further, let $n$-dimensional vectors,`$\varphi$' and `$\psi$', respectively denote the \textit{`cardinality'} and \textit{`feature'} learning set. For the $i^{th}$ particle, `$\beta_i$', and the corresponding learning exemplar, `$\alpha$', the learning sets are derived following the procedure outlined in Algorithm~\ref{alg:learningset}. This procedure is further explained by the illustrative example in \ref{s:appexample}. 

%------------------------------------------------------------------------
%------                     Pseudo Code:Position
%------------------------------------------------------------------------
\begin{algorithm}[!t]
    \small
    \SetKwInOut{Input}{Input}
    \SetKwInOut{Output}{Output}
    \SetKwComment{Comment}{*/ \ \ \ }{}
    \Input{$\mathcal{V}_i$}
    \Output{$\beta_i$}
    \BlankLine
    Set the new position $\beta_i$ to an $n$-dimensional null vector, \textit{i.e.}, $\beta_i=\{ 0 \dots 0 \}$ \\
    % \BlankLine
    Isolate the selection likelihood of the \emph{cardinality} and \emph{feature} into respective vectors, `$\rho$' and `$\sigma$' as follows: $\mathcal{V}_i = \begin{bmatrix} \rho \\ \sigma \end{bmatrix}$
    \BlankLine
    \Comment*[h] {Roulette Wheel Selection of the Subset Cardinality, ($\xi_i$)}\\
    Evaluate accumulative probabilities, $\rho_{\Sigma,j} = \sum \limits_{k=1}^{j} \rho_{k}, \ \  j = 1\dots n$. \nllabel{line:pos1}\\
    \BlankLine
    Generate a random number, $r \in [0,\rho_{\Sigma,n}]$. \nllabel{line:pos2}\\
    \BlankLine
    Determine the cardinality of the $i^{th}$ particle as follows: $\xi_i = \{j \ | \ \rho_{\Sigma,j-1}<r<\rho_{\Sigma,j} \}$
    % Determine `$j$' such that $\rho_{\Sigma,j-1}<r<\rho_{\Sigma,j}$, this gives the size of the subset $\xi_i$, \textit{i.e.}, $\xi_i=j$. \nllabel{line:pos3}\\
    \BlankLine
    \Comment*[h]{Selection of the features}\\
    Rank the features on the basis of their \emph{likelihood} `$\sigma_j$' and store the feature rankings in vector `$\tau$' \nllabel{line:fs1}\\
    \BlankLine
    \For{j = 1 to n} 
        { \If{$\tau_{j} \leq \xi_i$}
            {$\beta_{i,j}=1$}
            % {$\beta_{i,j}=0$}
        } \nllabel{line:fs2}
\caption{2-D learning approach to the position update of the $i^{th}$ particle}
\label{fig:posprop}
\end{algorithm}
%------------------------------------------------------------------------
%------------------------------------------------------------------------
%------                  Psuedo Code: UPSO
%------------------------------------------------------------------------
\begin{algorithm}[!ht]
    \small
    \SetKwInOut{Input}{Input}
    \SetKwInOut{Output}{Output}
    \SetKwComment{Comment}{*/ \ \ \ }{}
    \Input{\textit{PQ Dataset with $n$ features and wtih SNR$=\infty \ dB$}}
    \Output{\textit{Reduced Feature Subset `$\mathcal{X}$', $\xi_\mathcal{X} < n$}}
    \BlankLine
    Set the search parameters: $c_1, \ c_2, \  \omega, \ u, \ \& \  RG$ \\
    \BlankLine
    Randomly initialize the swarm of `$ps$' number of particles, $\{ \beta_1, \ \beta_2, \ \dots, \ \beta_{ps} \}$ \\
    Initialize the velocity ($(2 \times n)$ matrix) of each particle by uniformly distributed random numbers in [0,1] \\
    \BlankLine
    Evaluate the fitness of the swarm, $pbest$, $gbest$ and $nbest$ \\
    \BlankLine
    \For{t = 1 to iterations}
        {
            \BlankLine
            \Comment*[h]{Swarm Update}\\
            \BlankLine
            \For{i = 1 to ps} 
            {
                \BlankLine
                \Comment*[h]{Stagnation Check}\\
                \BlankLine
                \If{$ count_i \geq RG$}
                    {Re-initialize the velocity of the particle\\
                    Set $count_i$ to zero}
                \BlankLine
                Extract the learning sets,`$\mathcal{L}$', from each learning exemplar as per Algorithm~\ref{alg:learningset}\\
                Update the velocity of the $i^{th}$ particle as per~(\ref{eq:UPSOorig}), (\ref{eq:UPSOupdated1}) and~(\ref{eq:UPSOupdated2}) \\
                Update the position of the $i^{th}$ particle following Algorithm-\ref{fig:posprop}
            }
            \BlankLine
            Store the old fitness of the swarm in `$F$'\\
            \BlankLine
            Evaluate the swarm fitness\\
            \BlankLine
            Update personal, neighborhood and global best position, $pbest$, $nbest$ $gbest$\\             
            \BlankLine
            \Comment*[h]{Stagnation Check}\\ 
            \BlankLine
            \For{i = 1 to ps}
                { \If{$pbestval_i^{ \ t} \geq pbestval_i^{ \ t-1}$}
                        {$count_i=count_i+1$}
                }
        
        }
\caption{Pseudo code of 2D-UPSO algorithm for the feature selection problem}
\label{fig:2D-UPSO}
\end{algorithm}
%------------------------------------------------------------------------
%---------------------------------------------------------
\subsection{Velocity and Position Update}
\label{s:velupdate}

In this study, the well-known PSO variant, UPSO, has been adapted for the feature selection through the 2D-learning approach. The core idea behind UPSO \citep{UPSO1} is to combine the `\textit{global}' \citep{Shi:Eberhart:1998} and `\textit{local}' \citep{Kennedy:Mendes:2002} version of the PSO to achieve the balance between \textit{exploration} and \textit{exploitation} of the search landscape. The velocity update rule of UPSO adapted through 2D learning approach (referred as `2D-UPSO') is given by,
%------------------------------------------------------------------------
\begin{linenomath*}
\begin{align}
\label{eq:UPSOorig} 
\mathcal{V}_{i}  & = (u \times \mathcal{V}_{gi})+((1-u) \times \mathcal{V}_{li})\\
\label{eq:UPSOupdated1}
\text{where, \ } \mathcal{V}_{gi} & = (\omega \times \mathcal{V}_{i}) + (c_1r_1 \times \mathcal{L}_{cog}) + (c_2r_2 \times \mathcal{L}_{soc,1}) + (\Delta_{i} \times \mathcal{L}_{i}) \\ 
\label{eq:UPSOupdated2}
\mathcal{V}_{li} & = (\omega \times \mathcal{V}_{i}) + (c_1r_1 \times \mathcal{L}_{cog}) + (c_2r_2 \times \mathcal{L}_{soc,2}) + (\Delta_{i} \times \mathcal{L}_{i})
\end{align}
\end{linenomath*}
%------------------------------------------------------------------------
and `$u$' is the \textit{unification factor}; `$\omega$' is the \textit{inertia weight}; the parameters $[c_1, c_2]$ are \textit{acceleration constants}; $[r_1,r_2] \in [0,1]$ are \textit{uniform random numbers}; `$\mathcal{L}_{soc,1}$' and `$\mathcal{L}_{soc,2}$' are the \textit{social learning sets} derived respectively from global best ($gbest$) and neighbourhood best ($nbest_i$). Similarly, `$\mathcal{L}_{i}$' denotes the learning set derived from the current position of the particle, $\beta_i$. The adaptive weight used to control the influence of $\mathcal{L}_{i}$ is denoted as `$\Delta_i$' and it is given by,
%------------------------------------------------------------------------
\begin{linenomath*}
\begin{gather}
\label{eq:fitfeedback}
       \Delta_i= \begin{cases}
                + \delta_i, &   if \; \; \frac{f_{i}^{t}}{f_{i}^{t-1}}<1\\
                - \delta_i, &   otherwise
                 \end{cases} \ \ \text{where,\ }      \delta_i = 1- \frac{f_{i}^{t}}{max(F^{t})} %\nonumber
\end{gather}
\end{linenomath*}
%------------------------------------------------------------------------
where, `$f_{i}^{t}$' is the fitness of the $i^{th}$ particle and `$F^t$' is the vector which contains the fitness of the entire swarm at iteration `$t$'. For a given particle, the parameter `$\delta$' is \textit{adaptive} and its value is dependent on the relative performance of the particle with respect to the \textit{worst} particle of the swarm. As seen in (\ref{eq:fitfeedback}), the particle with the \textit{minimum} fitness will have \textit{higher} $\delta$, which will lead to the increase in selection likelihood of the corresponding features (included in the particle). Further, as seen in (\ref{eq:fitfeedback}), $\delta$ is positive only when the particle leads to improved fitness.

Following the velocity update, the new position of the particle is determined in two steps. In the first step, the cardinality of the new position is determined which is followed by the selection of the features. For this purpose, the \textit{selection likelihoods} stored in the velocity matrix are used. This procedure is outlined in Algorithm-\ref{fig:posprop} and further explained through an illustrative example in \ref{s:appexample}. The pseudo code for 2D-UPSO is outlined in Algorithm~\ref{fig:2D-UPSO}.

%----------------------------------------------------------------------
\begin{figure*}[!t]
\begin{minipage}[c]{.47\linewidth}
    \centering
    \includegraphics[width=\textwidth]{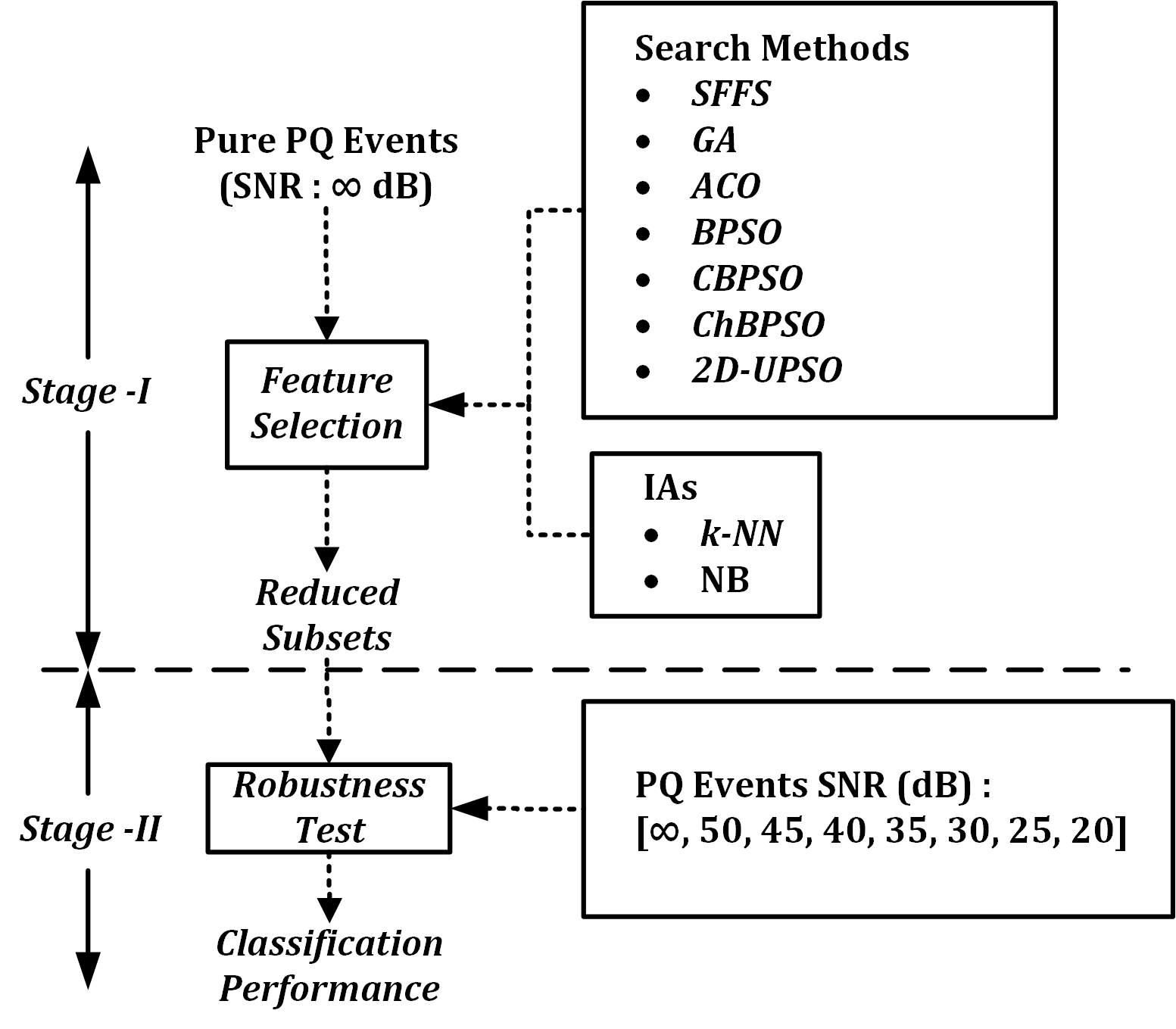}
    \caption{Investigation framework \label{f:if}}
\end{minipage}
\hfill
\begin{minipage}[c]{.5\linewidth}
\centering
\scalebox{0.99}{
\includegraphics [width=\textwidth]{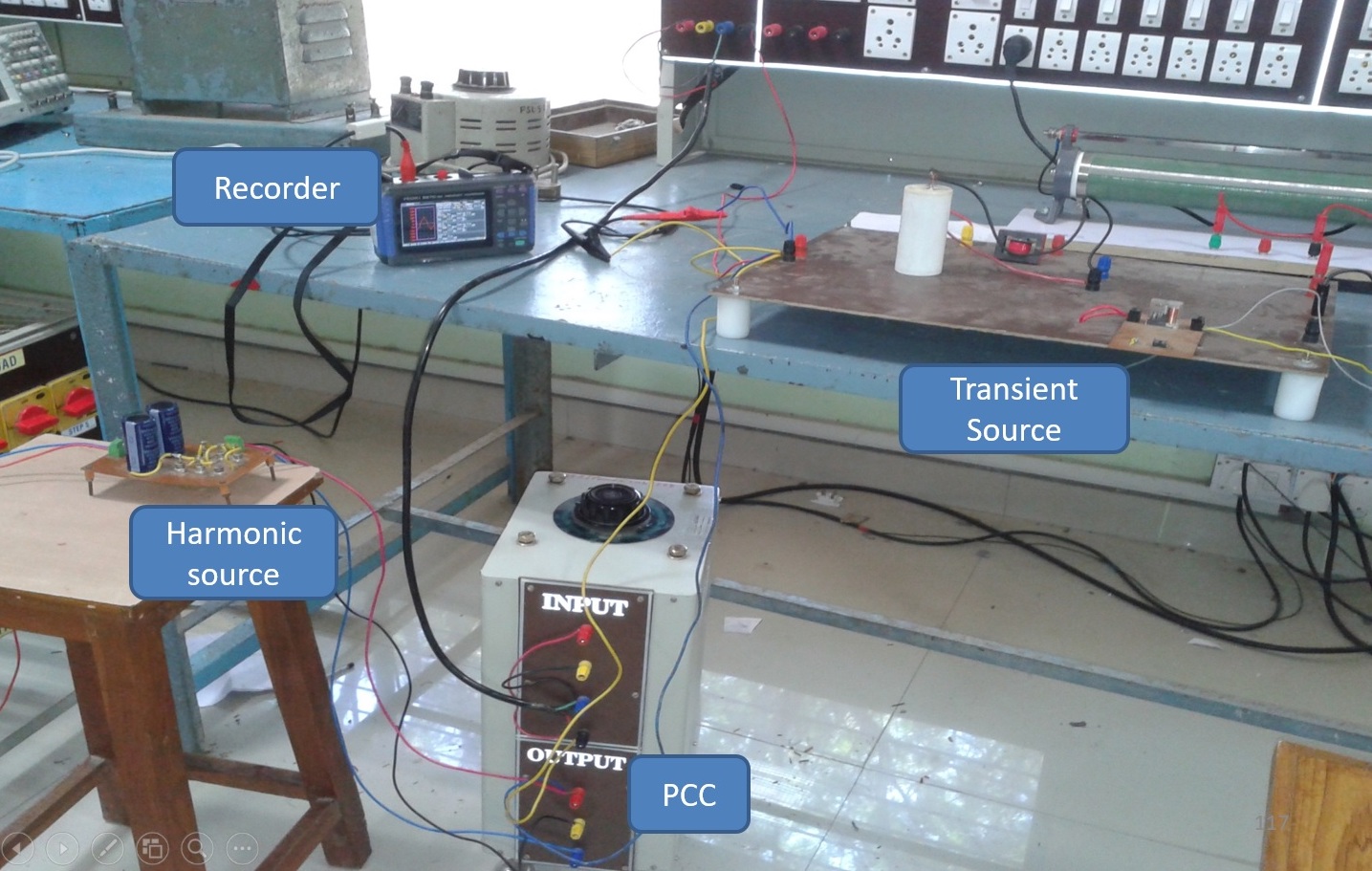}
}
\caption{The experimental setup to generate PQ events. The setup contains a three phase uncontrolled rectifier as a \textit{harmonic} source and a capacitor bank as a \textit{transient} source.}
\label{f:expsetup}
\end{minipage}
\end{figure*}%
%----------------------------------------------------------------------
%%%-------------------------- New Section ---------------------------------
\section{Investigation Framework}
\label{s:IF}

In this study, we focus mainly on two aspects of the feature selection in PQ events: 1) \textit{selection of an appropriate feature selection method} and 2) \textit{robustness of the reduced subsets against the measurement noise}. For this purpose, this investigation has been carried out in two stages, as shown in Fig.~\ref{f:if}. The objective of the first stage is to carry out the comparative analysis of different feature selection approaches. For this purpose, seven different search algorithms (discussed in Section~\ref{s:fsa}) are applied as a meta-heuristic wrapper to two induction algorithms (discussed in Section~\ref{s:IAPM}) and fourteen distinct types of PQ events (discussed in Section~\ref{s:PQE}). Note that, the PQ events used for the feature selection purpose do not contain any noise, \textit{i.e.}, Signal-to-Noise Ratio (SNR) $=\infty \ dB$. In the second stage, the robustness of the reduced subsets, obtained from the feature selection, is evaluated. For this purpose, the classification performance of the reduced subsets is evaluated in the presence of $7$ levels of zero mean Gaussian white noise: SNR=$[50, 45, 40, 35, 30, 25, 20] \ dB$. 

The following subsections provide more details about the PQ events, Search Algorithms and Induction Algorithms being used in this study.

%------------------------------------------------------------------------
%                       PQ Events
%------------------------------------------------------------------------
\begin{table*}[!t]
	    \caption{Power Quality Events$^\dagger$}
	    \label{t:events}
	    \begin{adjustbox}{max width=0.98\textwidth} 
	    \begin{threeparttable}
	   % \small 
	    \begin{tabular}{c c c} 
			\toprule
			\bfseries PQ Event & \bfseries Model & \bfseries Parameters$^{\dagger}$\\ [0.8ex]
			\midrule
			Pure Sinusoid & $ v_1(t) = \alpha \sin(\omega t)$ & $0.9<\alpha<1.1$\\ [1.0ex]
			\midrule	
			DC offset & $v_2(t) = \beta + \alpha \sin(\omega t)$ & $0.9<\alpha< 1.1$, $0 <\beta<0.1$\\ [1.0ex]
			\midrule
			Sag$^\ddagger$ & $v_3(t) = \{1-\alpha(u(t-t_1)-u(t-t_2))\} \times \sin(\omega t)$ & $0.1\leq\alpha\leq0.9$, $t_1<t_2$, $0.5T\leq t_2-t_1 \leq30T$ \\ [1.0ex]
			\midrule
			Swell$^\ddagger$ & $v_4(t) = \{1+\alpha(u(t-t_1)-u(t-t_2))\} \times \sin(\omega t)$ & $0.1\leq\alpha\leq0.8$, $t_1<t_2$, $0.5T\leq t_2-t_1 \leq30T$ \\ [1.0ex]
			\midrule
			Interruption$^\ddagger$ & $v_5(t) = \{1-\alpha(u(t-t_1)-u(t-t_2))\} \times \sin(\omega t)$ & $0.9 <\alpha\leq1.0$, $0.5T\leq t_2-t_1 \leq30T$ \\ [1.5ex]
			\midrule
			Flicker & $v_6(t) = \{1+\alpha_f \sin (\beta_f \omega t)\} \times \sin(\omega t)$ & \makecell{$ 0<\alpha_f\leq0.07$, $1\leq \beta_f \leq25$}\\ [1.5ex]
			\midrule
			Notching & \makecell{$v_7(t) = \sin(\omega t) - sign(\sin(\omega t)) \times $ \\  $\{\sum^{60}_{n=0}K [u(t-(t_1+nT))-u(t-(t_2+nT))] \}$} & \makecell{ $0.1\leq K\leq0.4$, $t_1 \leq T$ \\ $0.01T\leq t_2-t_1\leq 0.05T$} \\ [1.5ex]
		    \midrule
			Harmonics & $v_8(t) = \sum \limits_{k}^{} \delta_k \sin(\omega kt)$ & \makecell{$0< \delta_5,\delta_7,\delta_{11},\delta_{13}\leq 0.2$ \\ $\sum \limits_{k} \delta_k^2 = 1$, $k=\{ 1, 5, 7, 11, 13\}$} \\ [1.5ex]
			\midrule
			Oscillatory Transient & $v_9(t) = \sin(\omega t) + u_{t_1} \beta e^{-\gamma t_2} \sin(2\pi f_{tr}t)$ & \makecell{$50\leq\gamma\leq 100$, $1\leq\beta\leq 4$, $1000\leq f_{tr} \leq 10000$ \\ $0.3\leq t_1 \leq 0.9$, $t_2 =  (t-t_1)u_{t_1}$} \\  [2.0ex]
			\midrule
			Sag with Harmonics$^\ddagger$ & $v_{10}(t) = \{1-\alpha(u(t-t_1)-u(t-t_2))\} \times \sum \limits_{k}^{} \delta_k \sin(\omega kt)$ & \makecell{$ 0.5T \leq t_2-t_1 \leq 30T $, $ 0.1 \leq \alpha \leq 0.9 $, $0< \delta_5,\delta_7,\delta_{11},\delta_{13}\leq 0.2$ \\ $\sum \limits_k \delta_k^2 = 1$, $k=\{ 1, 5, 7, 11, 13\}$} \\ [1.8ex]
			\midrule
			Swell with Harmonics$^\ddagger$ & $v_{11}(t) = \{1+\alpha(u(t-t_1)-u(t-t_2))\} \times \sum \limits_{k}^{} \delta_k \sin(\omega kt)$ & \makecell{$ 0.5T \leq t_2-t_1 \leq 30T $, $ 0.1 \leq \alpha \leq 0.8 $, $0< \delta_5,\delta_7,\delta_{11},\delta_{13}\leq 0.2$ \\ $\sum \limits_k \delta_k^2 = 1$, $k=\{ 1, 5, 7, 11, 13\}$} \\  [1.8ex]
			\midrule
			Flicker with Harmonics & $v_{12}(t) = \{1+\alpha_f \sin(\beta_f \omega t)\} \times \sum \limits_{k}^{} \delta_k \sin(\omega kt)$ & \makecell{$0< \alpha_f \leq0.07$, $1 \leq \beta_f \leq 25$, $0< \delta_5,\delta_7,\delta_{11},\delta_{13}\leq 0.2$ \\ $\sum \limits_k \delta_k^2 = 1$, $k=\{ 1, 5, 7, 11, 13\}$} \\ [1.8ex]
			\midrule
			Sag with Transient$^\ddagger$ & \makecell{$v_{13}(t) = [\{1-\alpha(u(t-t_1)-u(t-t_2))\} \times  \sin(\omega t)] $ \\ $ + u_{t_1} \beta e^{-\gamma t_2} \sin(2 \pi f_{tr}t)$ }& \makecell{$ 0.1 \leq \alpha \leq 0.9 $, $50\leq\gamma\leq 100$, $1\leq\beta\leq 4$, $0.3\leq t_1 \leq 0.9$\\ $ 0.5T \leq t_2-t_1 \leq 30T $, $t_2 =  (t-t_1)u_{t_1}$ \\ $1000\leq f_{tr} \leq 10000$} \\  [1.5ex]
			\midrule
			Swell with Transient$^\ddagger$ & \makecell{$v_{14}(t) = [\{1+\alpha(u(t-t_1)-u(t-t_2))\} \times \sin(\omega t)] $ \\ $ + u_{t_1} \beta e^{-\gamma t_2} \sin(2 \pi f_{tr} t)$} & \makecell{$ 0.1 \leq \alpha \leq 0.8 $, $ 50 \leq \gamma \leq 100 $, $ 1 \leq \beta \leq 4 $, $0.3\leq t_1 \leq 0.9$ \\ $ 0.5T \leq t_2-t_1 \leq 30T $, $t_2 =  (t-t_1)u_{t_1}$ \\ $1000\leq f_{tr} \leq 10000$} \\ [1.5ex]
			\bottomrule
		\end{tabular}
		\begin{tablenotes}
        \item $^{\dagger}$ For each event duration $T = 30 \ cycles$; fundamental frequency $f = 50 \ Hz$; sampling frequency $F_s = 25 \ kHz$; 
        \item $^\ddagger$ $u(t)=1$ if $t>0$; otherwise $u(t)=0$
        \end{tablenotes}
        \end{threeparttable}
		\end{adjustbox}
\end{table*}
%------------------------------------------------------------------------
\subsection{PQ Events \& Feature Extraction}
\label{s:PQE}

For the comprehensive study, it is necessary to include a wide variety of PQ events in the investigation. For this purpose, the IEEE Std. 1159~\cite{IEEE:1159} is followed in this study. A total of fourteen distinct PQ events are generated through parametric models given in Table~\ref{t:events} and the experimental setup shown in Fig.~\ref{f:expsetup}. These events include several distinct natures of the PQ events, \textit{e.g.}, \textit{stationary} ($v_2, v_7, v_8$), \textit{non-stationary} ($v_3,v_4,v_5,v_9,\dots v_{14}$), \textit{low frequency} ($v_1, v_2, \dots v_7$), \textit{high frequency} ($v_8, v_9, \dots v_{14}$), \textit{single} ($v_1,v_2,\dots v_9$) and \textit{simultaneous} ($v_{10}, v_{11} \dots v_{14}$).

The synthetic PQ events were generated in MATLAB\textsuperscript{\textregistered} following the parametric models shown in Table~\ref{t:events}. The real PQ events are acquired using the experimental PCC shown in Fig.~\ref{f:expsetup}, which consist of a harmonic source and a transient source. The harmonic source is being emulated by a three-phase uncontrolled rectifier with a resistive load which generates the harmonics of order $(6k \pm 1), \ k=\{ 1, 2, 3, \dots \}$. The transient events are induced by switching of a capacitor bank. Further, the sag events are induced by creating a single line to ground fault. A digital oscilloscope/recorder (HIOKI 8870-20 MEMORY HiCORDER\textsuperscript{\textregistered}) is used to capture the real events. For each class of PQ event, $250$ instances are generated which gives a total of $3500$ instances of PQ events. Approximately $90\%$ of these event instances are generated through the parametric models shown in Table~\ref{t:events} and the remaining are induced using the experimental PCC shown in Fig.~\ref{f:expsetup}. Each event instance is generated at the fundamental frequency of $50 \ Hz$, for the duration of $30$ cycles and sampled at $25 \ kHz$. The PQ events include magnitude variation in the range of $[0,4] \ pu$, frequency variation in the range of $[0,10] \ kHz$, harmonics of order $\{ 5, 7, 11, 13, 17, 19, 23, 25, 29, 31, \dots \}$  and the event duration from $0.5$ to $30$ \textit{cycles}. In addition, to evaluate the effects of the measurement noise, seven different levels of zero-mean Gaussian white noise have been added to these events. This gives a total of $8$ PQ datasets; each dataset has the same $3500$ events but a different level of the measurement noise, \textit{i.e.}, SNR$=[ \infty, 50, 45, 40, 35, 30, 25, 20] \ dB$.

For accurate identification of the PQ events, it is crucial to extract information in both temporal and frequency domain. This task can be accomplished through various signal processing techniques such as Stockwell-Transform (ST), Wavelet Packet Transform (WPT), Discrete Wavelet Transform (DWT)~\cite{Mahela:Shaik:2015,Khokhar:Zin:2015}. Note that the selection of signal processing technique is primarily dependent on the frequency bandwidth of the signal being investigated. Especially, a judicious selection is essential to accommodate the relatively large frequency bandwidth of PQ events, \textit{e.g.}, from \textit{DC offset} ($0 \ Hz$) to \textit{Oscillatory Transients} ($5 \ MHz$)~\cite{IEEE:1159}. In this scenario, among existing signal processing techniques, DWT represents an ideal choice for PQ events as it is computationally more efficient; the complexity of DWT, WPT and ST is respectively $\mathcal{O}(N)$, $\mathcal{O}(N log N)$, $\mathcal{O}(N^2 log N)$ (\textit{$N$ denotes the number of samples}). Therefore, DWT has been selected as a signal processing technique in this study. 

It is well-known that the selection of the \textit{base/mother} wavelet is dependent on the nature of the application and it is crucial to the performance of wavelet transforms. The detailed investigation on this topic~\cite{Hafiz:Swain:2019} suggest that for the PQ events, the optimum classification performance of the given induction algorithm can be obtained when it is paired with a specific base wavelet. For example, the optimum performance from \textit{k-Nearest Neighbor} (k-NN) and \textit{Naive-Bayes} (NB) (\textit{which are being used in this study}) was obtained with the $6^{th}$ order \textit{symlet} (`$sym6$')~\cite{Hafiz:Swain:2019}. Therefore, in this study, `$sym6$' has been selected as the \textit{base wavelet}.

Further, each instance of the PQ event is decomposed to the $8^{th}$ level using DWT (\textit{the decomposition level has been selected following the rule of thumb given in~\ref{s:decomplvl}}). Consequently, the \textit{detail coefficients} at each level and \textit{approximation coefficients} at the final level are available. In order to extract meaningful information from the wavelet coefficients, $11$ \textit{statistical functions} are used (\textit{shown in}~\ref{s:app}). Following this procedure, for each instance of PQ event, a total of $99$ \textit{`features'} ($9$ \textit{details/approximations} $\times$ $11$ \textit{functions}) are obtained. Hence, a \textit{pattern}, `$\ell$', is obtained corresponding to each instance of PQ event, as follows: 
%----------------------------------------------------------------------
\begin{equation}
\label{eq:U}
\ell =\{ u_1,u_2,\dots u_{n}, v_k \}, \quad v_k \in V
\end{equation}
%------------------------------------------------------------------------
where, `$n=99$' denotes the \textit{number of features}; $U=\{ u_1,u_2,\dots u_{n} \}$  denotes the \textit{features set} and $V=\{ v_1,v_2,\dots v_{14} \}$ contains the \textit{label} corresponding to each PQ event. 

%------------------------------------------------------------------------
\subsection{Induction Algorithm}
\label{s:IAPM}

The results of our previous investigation~\cite{Hafiz:Swain:2019} suggest that relatively simple induction algorithms are more robust against the measurement noise. In particular, \textit{Decision Tree} (DT) and \textit{Naive-Bayes} (NB)~\cite{Witten:Frank:Hall:2016} were found to be comparatively more robust for PQ events. Since DT has limited inherent feature selection capability, we have selected NB as one of the induction algorithms in this study. Further, it is well-known that the search landscape of the feature selection problem is conjointly defined by the dataset and the induction algorithm~\cite{Blum:Langley:1997,Dash:Liu:1997}. For this reason, in addition to NB, \textit{k-Nearest Neighbor} (k-NN) \cite{Witten:Frank:Hall:2016} is also used as an induction algorithm in this study.

%------------------------------------------------------------------------
\begin{figure*}[!t]
\centering
\begin{subfigure}{.42\textwidth}
  \centering
  \includegraphics[width=\textwidth]{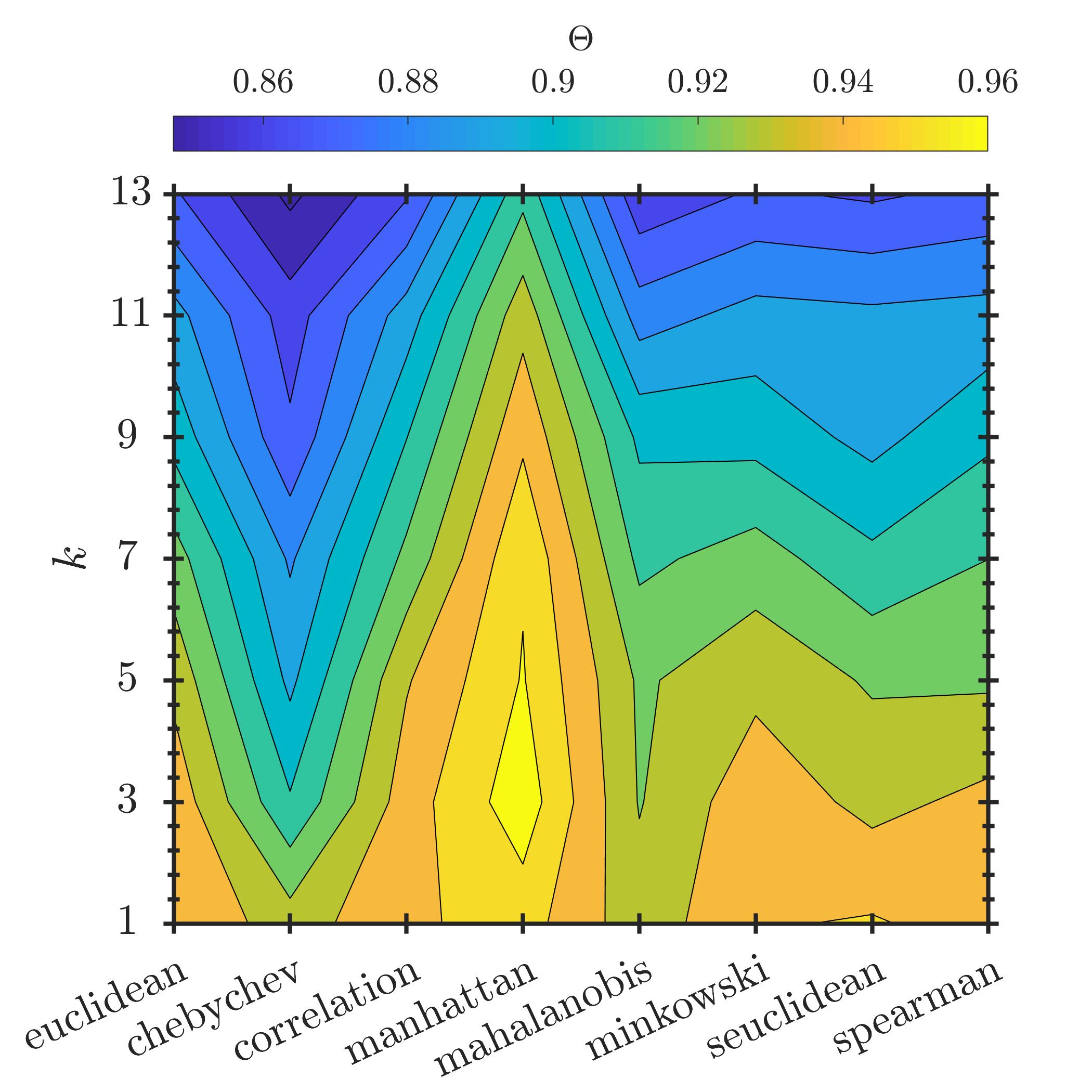}
  \caption{Selection of `$k$' and distance metric in k-NN. The maximum $\Theta$ is obtained with `$k=3$' and \textit{`Manhattan'} distance.}
  \label{f:knnpara}
\end{subfigure}
\hfill
\begin{subfigure}{.43\textwidth}
  \centering
  \includegraphics[width=\textwidth]{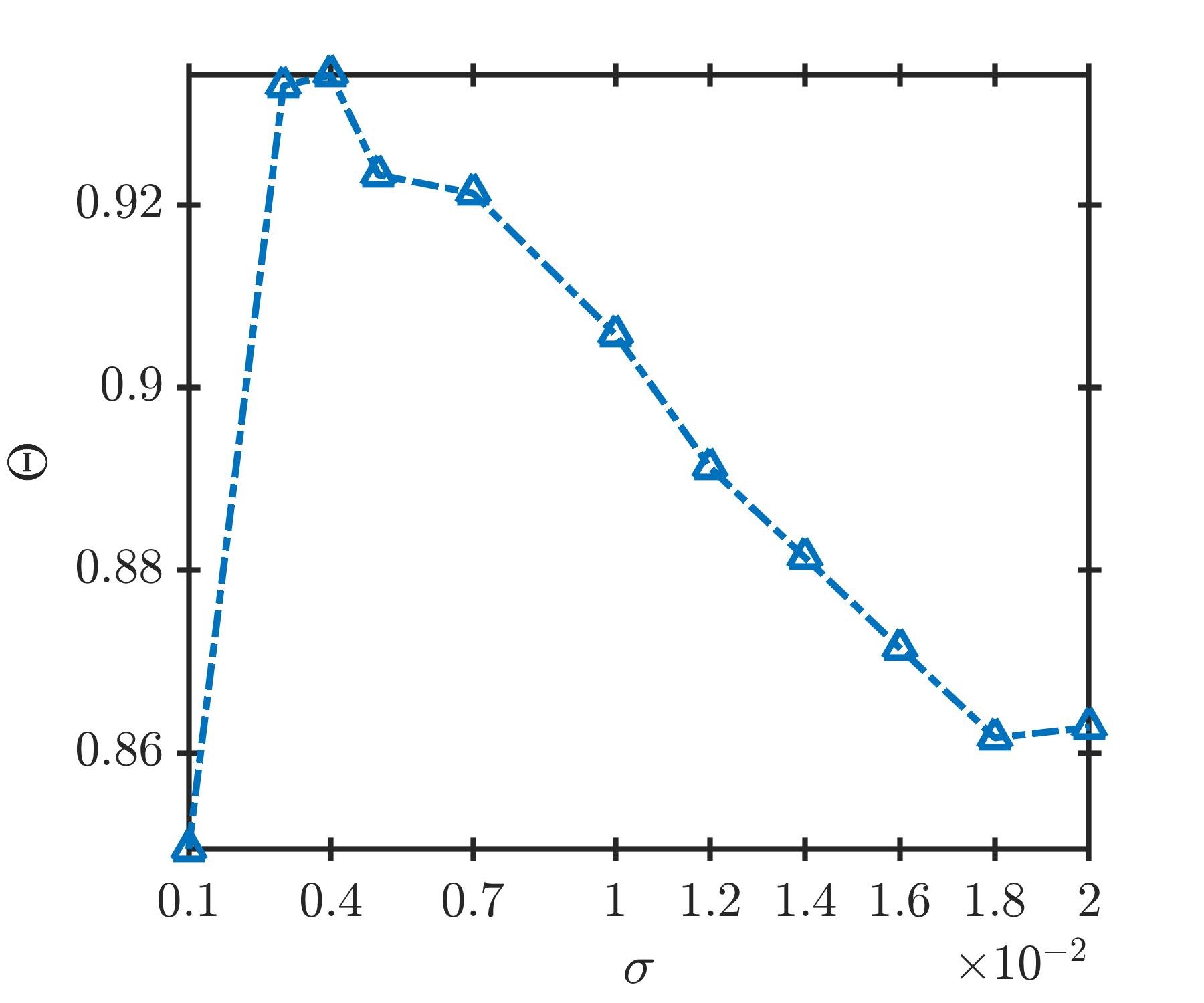}
  \caption{Selection of kernel width `$\sigma$' in NB. The maximum $\Theta$ is obtained with `$\sigma=0.004$'.}
  \label{f:nbpara}
\end{subfigure}
\caption{Grid-search for Hyperparameter selection. `$\Theta$' denotes the \textit{average} ten-fold classification accuracy.}
\label{f:hyper}
\end{figure*}
%------------------------------------------------------------------------

Note that the classification performance of IA is critically dependent on the `hyperparameters' which are used to control the learning process, \textit{e.g.}, \textit{`number of neighbors'} ($k$) and \textit{`distance metric'} in k-NN; `\textit{kernel width}' ($\sigma$) in NB. In this study, the hyperparameters of both IAs have been selected through \textit{`grid-search'} to maximize the average ten-fold classification accuracy (`$\Theta$'), as shown in Fig.~\ref{f:knnpara} (k-NN) and Fig.~\ref{f:nbpara} (NB).

\subsection{Compared Feature Selection Algorithms}
\label{s:fsa}

To evaluate the efficacy of the 2D learning approach, the following existing algorithms are considered: Genetic Algorithm (GA)~\cite{Siedlecki:Sklansky:1989,Kudo:Sklansky:2000}, Ant Colony Optimization (ACO) \cite{Yu:Gu:2009,Chen:Chen:2013}, Binary PSO (BPSO)~\cite{Kennedy:Eberhart:1997}, Catfish BPSO~\cite{Chuang:Tsai:2011} and Chaotic BPSO (chBPSO)~\cite{Chuang:Yang:2011}. In all these algorithms, the search agent (\textit{e.g.}, a \textit{chromosome} in GA) encodes a feature subset by the binary string representation given in~(\ref{eq:binaryrepresentation}).

In this study, the classical variant of GA is considered, \textit{i.e.}, simple GA with \textit{roulette wheel} selection, \textit{single-point} crossover and \textit{flip-bit} mutation operator, as outlined in~\cite{Siedlecki:Sklansky:1989,Kudo:Sklansky:2000}. For ACO, the variant proposed in~\cite{Yu:Gu:2009,Chen:Chen:2013} is considered, where the features are represented as \textit{nodes} on a directed graph. Each node on this graph is linked by two distinct \textit{edges} to highlight whether a node/feature is selected or not. A feature subset is represented by a \textit{path} traversed by an \textit{ant} over these edges. For the given \textit{edge}, the probability of inclusion in the \textit{path} is given by the corresponding \textit{pheromone intensity}. In each iteration, the pheromone intensity is updated based on the positive feedback mechanism. In this study, the pheromone update procedure proposed for ACO in~\cite{Yu:Gu:2009} is implemented. Further, since the proposed 2D-UPSO has been developed in particle swarm theory, BPSO and its two variants (CBPSO and chBPSO) are also included in the comparative investigation. The CBPSO~\cite{Chuang:Tsai:2011} retains the learning mechanism of BPSO while introducing the concept of \textit{`refresh gap'}, \textit{i.e.}, a fixed number of worst performing particles are reinitialized if the swarm cannot locate an improved solution for a pre-fixed number of iterations. In chBPSO~\cite{Chuang:Yang:2011}, the velocity update rule of BPSO is updated to control the inertia weight following the chaotic maps. In this study, a \textit{logistic map} is being used to determine the value of inertia weight in chBPSO.  

In addition to the meta-heuristic algorithms, the Sequential Forward Floating Search (SFFS)~\cite{Pudil:1994} is also included to compare the search performance of the 2D learning with the existing deterministic search algorithms. The SFFS has been selected for this purpose, as it overcomes the limitations of the other sequential search methods~\cite{Pudil:1994}, \textit{e.g.}, \textit{nesting effects} of SFS and SBS; selection of appropriate `$(l,r)$' in \textit{plus-l take away-r floating search}.

The search parameters of each algorithm are set following the procedures outlined in \cite{Siedlecki:Sklansky:1989,Yu:Gu:2009,Kennedy:Eberhart:1997,Chuang:Tsai:2011,Chuang:Yang:2011} and are shown in Table \ref{t:sparam}. All of the compared algorithms are implemented in MATLAB.

%---------------- Table: Parameter Settings-------------------------
\begin{table}[!t]
\centering
\small
\caption{Search Parameter Settings}
\label{t:sparam}
\begin{adjustbox}{max width=0.9\textwidth}
\begin{threeparttable}
\begin{tabular}{c c c c c c} 
 \toprule
  \multirow{3}{*}{\textbf{Algorithm}} &  \multicolumn{2}{c} {\textbf{Search Parameters}}  \\ [1ex]
   & \textbf{General PSO Parameters} & \multirow{2}{*}{\textbf{Other/Special Parameters}} \\ [1ex]
   & \boldmath$[ps, \ \omega, \ c_1, \ c_2, \ v_{min}, \ v_{max} ]$  &   \\ [1.0ex]
 \midrule
GA \cite{Siedlecki:Sklansky:1989} &  - &  $N=30$, $p_c=0.8$, $p_m=0.2$\\ [1.5ex]
ACO \cite{Yu:Gu:2009} &  - &   \makecell{$as=50$, $a=5$, $\rho=0.2$, \\ $[\tau_{min},\tau_{max}]=[0.3, 1.5]$}\\ [1.9ex]
BPSO \cite{Kennedy:Eberhart:1997}                   &  $[30, \ 1, \ 2, \ 2, \ -6, \ 6]$ & - \\ [0.8ex]
CBPSO \cite{Chuang:Tsai:2011}                &  $[30, \ 1, \ 2, \ 2, \ -6, \ 6]$ & $C=\frac{ps}{10}$, RG=3 \\ [0.8ex]
chBPSO \cite{Chuang:Yang:2011}                     &  $[30, -, \ 2, \ 2, \ -6, \ 6]$ & variable $\omega$ \\ [0.8ex]
2D-UPSO                                   &  $[30, \ 1, \ 2, \ 2, \ -, \ -]$ & $u:0.9$, $RG=30$\\ [0.8ex]
\bottomrule
\end{tabular}
\begin{tablenotes}
      \small
      \item `$ps$' - swarm size, `$\omega$' - inertia weight, `$c_1,c_2$' - acceleration constants; $[v_{min},v_{max}]$ - velocity limits, $[x_{min},x_{max}]$ - position limits; 
      \item`$N$' - GA population size, $p_c,p_m$ - crossover and mutation probability
      \item `$as$' - colony size, `$a$' - pheromone update factor, `$\rho$' - pheromone trail evaporation, $[\tau_{min},\tau_{max}]$ - pheromone boundaries  
      \item `$C$' - catfish particles, `$u$' - unification factor
    \end{tablenotes}
  \end{threeparttable}
 \end{adjustbox}
\end{table}
%---------------------------------------------------------

\subsection{Search Setup}
\label{s:searchsetup}

The efficacy of the feature subset can be evaluated by either the \textit{filter} or the \textit{wrapper} approach. The selection of either approach requires a trade-off between precision and the computational complexity. In this study, the total number of features is moderate ($n=99$), and therefore it is appropriate to select the \textit{wrapper} approach due to its precision \cite{Blum:Langley:1997}. In other words, for each subset under consideration, a classifier is induced by the induction algorithm (\textit{say NB or k-NN}) and the subsequent classification error is used as the criterion function, $J(\cdotp)$.

Further, to remove any bias to the validation data, the mean classification error after \textit{10-fold stratified cross-validation} is used as the criterion function, $J(\cdotp)$ \cite{Witten:Frank:Hall:2016}. For a given feature subset, $\mathcal{X}$, this is given by,
%---------------------------------------------------------
\begin{linenomath*}
\begin{align}
    \label{eq:fitness}
        J(\mathcal{X}) & = \frac{1}{10} \times \sum \limits_{f=1}^{10} e_f, \\
        \text{where, } e_f & = \frac{\textit{mis-classified events in the $f^{th}$-fold}}{\textit{total number of events in the $f^{th}$-fold}} \nonumber
\end{align}
\end{linenomath*}
%---------------------------------------------------------
Without loss of generality, the feature selection problem is approached as a \textit{minimization} problem where the objective is to minimize the classification error, $J(\cdotp)$ given by (\ref{eq:fitness}). To account for the inherent stochastic nature of the algorithms, $40$ independent runs of each algorithm are executed. Each run is set to terminate after $6000$ Function Evaluations (FEs).

%-------------------------- New Section ---------------------------------
\section{Results}
\label{s:res}

As mentioned earlier, in this study we address the following two issues: 1) \textit{Which feature selection approach is more effective for PQ events?} and 2) \textit{How robust are the reduced feature subsets against the measurement noise?} For this purpose, the comparative evaluation of seven feature selection wrappers has been carried out using two induction algorithms considering fourteen distinct PQ events, following the procedure outlined in `Stage-I' of Fig.~\ref{f:if}. Note that the feature selection has been carried out using only the `\textit{pure}' PQ events, \textit{i.e.}, with SNR$=\infty \ dB$. The results of this investigation have been discussed in Section~\ref{s:ce}. 

The next issue is to evaluate the robustness of the reduced feature subsets which are obtained by the feature selection algorithms. For this purpose, the following seven different levels of zero mean Gaussian white noise are added to the PQ events: SNR$=[50, 45, 40, 35, 30, 25, 20] \ dB$. The framework for this part of the investigation is outlined in `Stage-II' of Fig.~\ref{f:if}. The results of this test are discussed in Section~\ref{s:test}.

%----------------------------------------------------------------------
%--                      Table - Search Results
%----------------------------------------------------------------------
\begin{table*}[!t]
%  \begin{minipage}[c]{.5\linewidth}
    \centering
    \caption{Search performance of the compared algorithms with k-NN (averaged over 40 runs)} \label{t:res:knn}
    \begin{adjustbox}{max width=0.85\textwidth}
    \small
    \begin{threeparttable}
    \small
    \begin{tabular}{cccccccc}
    \toprule
    \multicolumn{2}{c}{\textbf{Results}} & \textbf{GA} & \textbf{ACO} & \textbf{BPSO} & \textbf{CBPSO} & \textbf{chBPSO} & \textbf{2D-UPSO} \\ [1.2ex]
    \midrule
    
    \multirow{4}{*}{$J(\cdotp)$} & $Mean$  & 0.0161 & 0.0147 & 0.00552 & 0.00558 & 0.0094 & 0.0044 \\[1ex]
          & $SD$    & $1.1\times10^{-3}$ & $1.3\times10^{-3}$ & $4.6\times10^{-4}$ & $5.3\times10^{-4}$ & $7.6\times10^{-4}$ & $1.1\times10^{-3}$ \\[0.8ex]
          & $PI(\%)$ & 53.1  & 57.1  & 83.9  & 83.7  & 72.5  & \textbf{87.0} \\[0.8ex]
        %   & Rank  & 6     & 5     & 2     & 3     & 4     & 1 \\[0.8ex]
    \midrule
    \multirow{4}{*}{$\xi$} & $\xi_{avg}$ & 47.6  & 45.7  & 39.8  & 38.7  & 41.1  & 20.9 \\[0.8ex]
          & $SD$    & 5.0 & 5.3 & 4.3 & 4.9 & 4.4 & 7.0 \\[0.8ex]
          & $\Xi(\%)$ & 52.0  & 53.9  & 59.8  & 60.9  & 58.5  & \textbf{78.9} \\[0.8ex]
        %   & Rank  & 6     & 5     & 3     & 2     & 4     & 1 \\[0.8ex]
    \midrule
    \textit{Overall Score} &       & 0.3096 & 0.2719 & 0.0892 & 0.0879 & 0.1571 & \textbf{0.0399} \\%[0.8ex]
    % \textit{Overall Rank} &       & 6     & 5     & 3     & 2     & 4     & 1 \\

    \bottomrule
    
    \end{tabular}%
    \begin{tablenotes}
      \scriptsize
      \item `Mean'and 'SD' - Mean and standard deviation over 40 runs
      \item `PI(\%)'- Improvement in the classification error relative to the original feature set, $J(U)=0.0343$
      \item `$\xi_{avg}$' - Average length of the feature subset over 40 runs
      \item $\Xi(\%)$ - Percentage reduction in the subset size
    \end{tablenotes}
    \end{threeparttable}
    \end{adjustbox}
%  \end{minipage}
 \end{table*}
%  \hfill
%   \begin{minipage}[c]{.5\linewidth}
\begin{table*}[!t]
    \centering
    \caption{Search performance of the compared algorithms with NB (averaged over 40 runs)} \label{t:res:nb}
    \begin{adjustbox}{max width=0.85\textwidth}
    \small
    \begin{threeparttable}
    \small
    \begin{tabular}{cccccccc}
    \toprule
    \multicolumn{2}{c}{\textbf{Results}} & \textbf{GA} & \textbf{ACO} & \textbf{BPSO} & \textbf{CBPSO} & \textbf{chBPSO} & \textbf{2D-UPSO} \\
    \midrule
   
    \multirow{4}{*}{$J(\cdotp)$} & $Mean$  & 0.0136 & 0.0128 & 0.00663 & 0.00652 & 0.0095 & 0.0061 \\[1ex]
          & $SD$    & $7.7\times10^{-4}$ & $8.5\times10^{-4}$ & $1.1\times10^{-3}$ & $3.3\times10^{-4}$ & $5.9\times10^{-4}$ & $2.5\times10^{-4}$ \\[1ex]
          & $PI(\%)$ & 31.8  & 35.6  & 66.7  & 67.3  & 52.4  & \textbf{69.4} \\[1ex]
        %   & Rank  & 6     & 5     & 3     & 2     & 4     & 1 \\[1ex]
    \midrule
    \multirow{4}{*}{$\xi$} & $\xi_{avg}$ & 49.3  & 49.0  & 43.0  & 43.8  & 46.3  & 37.0 \\[0.8ex]
          & $SD$    & 3.8 & 4.8 & 4.2 & 3.8 & 4.0 & 4.4 \\[1ex]
          & $\Xi(\%)$ & 50.3  & 50.6  & 56.6  & 55.8  & 53.3  & \textbf{62.6} \\[1ex]
        %   & Rank  & 6     & 5     & 2     & 3     & 4     & 1 \\[1ex]
    \midrule
    \textit{Overall Score} &       & 0.2707 & 0.2546 & 0.1156 & 0.1156 & 0.1776 & \textbf{0.0913} \\%[0.8ex]
    % \textit{Overall Rank} &       & 6     & 5     & 2     & 3     & 4     & 1 \\

    \bottomrule
    
    \end{tabular}%
    \begin{tablenotes}
      \scriptsize
       \item `Mean'and 'SD' - Mean and standard deviation of classification error, $J(\mathcal{X})$, over 40 runs
      \item `PI(\%)'- Improvement in the classification error relative to the original feature set, $J(U)=0.0199$ 
      \item `$\xi_{avg}$' - Average length of the feature subset over 40 runs
      \item $\Xi(\%)$ - Percentage reduction in the subset size
    \end{tablenotes}
    \end{threeparttable}
    \end{adjustbox}
%  \end{minipage}
 \end{table*}
%----------------------------------------------------------------------

\subsection{Stage-I : Comparative evaluation of the feature selection approaches}
\label{s:ce}

For the purpose of comparative evaluation, $40$ independent runs of each algorithm are recorded. Since the primary objective is to improve the classification performance through the removal of irrelevant/redundant features, the search performance of the compared algorithms is evaluated by two criteria, \textit{i.e.}, \textit{classification performance} and \textit{size of the feature subset (cardinality)}. 

The results obtained after 40 independent runs of each algorithm with k-NN and NB classifier are shown in Table~\ref{t:res:knn} and~\ref{t:res:nb}, respectively. The results obtained with 2D learning approach (2D-UPSO) are shown in the last column of Table~\ref{t:res:knn} and~\ref{t:res:nb}. The best results obtained among the compared algorithms are shown in boldface.

To compare the \textit{classification performance}, the \textit{average} (Mean) and \textit{standard deviation} (SD) of the criterion function, $J(\cdotp)$, are shown in Table \ref{t:res:knn} and \ref{t:res:nb}. Similarly, to compare the reduction in subset size, `Mean' and `SD' of subset cardinality is computed over 40 runs for each algorithm as shown in Table \ref{t:res:knn} and \ref{t:res:nb}. Further, the following two metrics are used to measure the performance improvement obtained by each of the algorithms,
%----------------------------------------------------------------------
\begin{linenomath*}
\begin{align}
    PI (\%) & =\frac {J(U)-\overline{J(\mathcal{X})}}{J(U)} \times 100\\
    \Xi (\%) & = \frac {n - \xi_{avg}}{n} \times 100
\end{align}
\end{linenomath*}
%----------------------------------------------------------------------
where, `$J(U)$' is the criterion function with the original feature set ($U$) and `$\overline{J(\mathcal{X})}$' is the average of the criterion function over 40 runs obtained with the reduced feature sets; `$\xi_{avg}$' is the average cardinality over 40 runs and `$n$' is the total number of features, \textit{i.e.}, $n=99$.

The \textit{Performance Improvement}, $PI(\%)$, metric gives the improvement in the classification performance with respect to the original feature set, `$U$'. The second metric, $\Xi(\%)$, shows the \textit{percentage reduction in the cardinality} with respect to the total number of features, `$n$'. A higher positive value of these metrics implies a better search performance. In addition, the following metric is used to evaluate the overall performance of the algorithms:
%----------------------------------------------------------------------
\begin{equation}
    \label{eq:score}
    \textit{overall score} = \sum \limits_{k=1}^{40} (\frac{\xi_k}{n}) \times J(\mathcal{X}_k) %\frac{1}{40}\sum \limits_{k=1}^{40} \{\frac{n-\xi_k}{n}\} \times \{1-J(X_k)\}
\end{equation}
%----------------------------------------------------------------------
where, `$\mathcal{X}_k$', `$\xi_k$' and `$J(\mathcal{X}_k)$' are the feature subset, its cardinality and the corresponding classification error at the end of `$k^{th}$' run of the algorithm; `$n$' denotes the total number of features. Note that this metric incorporates the information about both the cardinality and the classification performance. A \textit{lower} value of this metric indicates that the algorithm could consistently find feature subsets with a \textit{lower cardinality} and a \textit{lower classification error}.

From the results shown in Table \ref{t:res:knn} and \ref{t:res:nb}, it is clear that the 2D learning approach is most efficient as it gives the lowest classification error amongst the compared algorithms. With both the induction algorithms, 2D-UPSO could achieve the highest $PI(\%)$, approximately $87\%$ (with k-NN, Table \ref{t:res:knn}) and $69\%$ (with NB, Table \ref{t:res:nb}). It is interesting to note that in comparison to GA and ACO, all BPSO variants could yield better subsets with both k-NN and NB.

Further, in the present study, the feature selection issue is approached as a \textit{single-objective} problem where the primary objective is to minimize the classification error. Therefore, any reduction in the subset cardinality is the direct consequence of the ability of the search algorithms to distinguish useful features from the irrelevant/redundant features. Intuitively, the exploitation of cardinality information in the \textit{2D learning} is likely to improve the search performance of 2D-UPSO. The results shown in Table \ref{t:res:knn}-\ref{t:res:nb} corroborates this notion. 2D-UPSO could provide the smallest feature subset with the highest reduction in the cardinality,\textit{e.g.}, $78.9\%$ (with k-NN) and $62.6\%$ (with NB).

The overall performance of the compared algorithms is evaluated using the `\textit{overall-score}' metric (\ref{eq:score}), which considers both the cardinality, $\xi$, and the classification performance, $J(\cdotp)$, of the feature subsets obtained by the algorithm over 40 runs. As revealed by (\ref{eq:score}), a lower \textit{score} indicates the consistent discovery of a subset with \textit{fewer features} and \textit{lower classification error} by the algorithm. As seen in Table \ref{t:res:knn}-\ref{t:res:nb}, the overall score obtained by 2D-UPSO is the lowest amongst the compared algorithms which indicates the best overall performance.

The results further show a shift in the search landscape with the change of induction algorithms. For example, with k-NN, $PI(\%)$ from the compared algorithms lies in the range of $53\%-87\%$ (Table \ref{t:res:knn}) whereas with NB the performance gain is comparatively lower (in the range of $32\%-69\%$, Table \ref{t:res:nb}). Similar effects are observed in $\Xi(\%)$ as well; its variation with k-NN and NB classifier lie in the range of $52\%-79\%$ and $50\%-62\%$, respectively. These results further underline the need for the wrapper based feature selection approach.

%---------------------------------------------------------
\subsubsection{Nonparametric Statistical Evaluation}
\label{s:Fried}

Due to the stochastic nature of the compared algorithms, further statistical analysis is carried out to determine the significance of the results shown in Table~\ref{t:res:knn} and Table~\ref{t:res:nb}. In particular, the objective of this analysis is to determine whether the results (\textit{i.e.}, $J(\cdotp)$ and $\xi$) obtained by 2D-UPSO, are significantly better than the compared algorithms. For this purpose, multiple non-parametric statistical comparisons are carried out following the guidelines in Derrac \etal~\cite{Derrac:Salvador:2011}. The test is carried out in the following two steps:

%---------------------------------------------------------
\begin{table}[!t]
  \centering
  \small
  \caption{Outcome of the Friedman Test}
  \label{t:Fr1}%
  \begin{adjustbox}{max width=0.88\textwidth}
  \begin{threeparttable}
  
   \begin{tabular}{cccccccccc}
    \toprule
    \multicolumn{2}{c}{\multirow{2}[4]{*}{\textbf{Results}}} & \multicolumn{6}{c}{\textbf{Average Rank}$^\dagger$}     & \multirow{2}{*}{\makecell{\textbf{Friedman} \\ \textbf{Statistic}}} & \multirow{2}{*}{\textit{\textbf{p-value}}} \\
    \cmidrule{3-8}    \multicolumn{2}{c}{} & \textbf{GA} & \textbf{ACO} & \textbf{BPSO} & \textbf{CBPSO} & \textbf{chBPSO} & \textbf{2D-UPSO} &       &  \\[1ex]
    \midrule
    \multirow{2}{*}{k-NN} & $J(\cdotp)$ & 5.9   & 5.2   & 2.3   & 2.4   & 4.0   & \textbf{1.3}   & 182.66 & 1.23E-10 \\[1ex]
          & $\xi$ & 5.2   & 4.9   & 3.3   & 2.9   & 3.7   & \textbf{1.0}   & 130.25 & 7.95E-11 \\[1ex]
    \midrule
    \multirow{2}{*}{NB} & $J(\cdotp)$ & 5.7   & 5.3   & 2.3   & 2.5   & 4.0   & \textbf{1.2}   & 181.58 & 8.16E-11 \\[1ex]
          & $\xi$ & 4.9   & 4.8   & 2.9   & 3.2   & 3.9   & \textbf{1.3}   & 101.06 & 9.26E-11 \\[1ex]
    \bottomrule
    \end{tabular}%
    \begin{tablenotes}
      \scriptsize
       \item $^\dagger$ the best average ranking is shown in bold-face
    \end{tablenotes}
   \end{threeparttable}
   \end{adjustbox}    
\end{table}%
%---------------------------------------------------------
%---------------------------------------------------------
\begin{table}[!t]
  \centering
  \small
  \caption{Outcome of the Hommel's Post-hoc Procedure for $95\%$ Confidence Interval (k-NN)}
  \label{t:Fr2}%
  \begin{adjustbox}{max width=0.8\textwidth}
  \begin{threeparttable}
    \begin{tabular}{ccccc|cccc}
    \toprule
    
    \multirow{2}{*}{\textbf{Algorithm}} & \multicolumn{4}{c}{\boldmath$J(\cdotp)$} & \multicolumn{4}{c}{\boldmath$\xi$} \\
    \cmidrule{2-9}  & \makecell{\textbf{Test} \\ \textbf{statistic}}     & \boldmath$p-$\textbf{value} & \textbf{APV} & \boldmath$H_0^\dagger$  & \makecell{\textbf{Test} \\ \textbf{statistic}}     & \boldmath$p-$\textbf{value} & \textbf{APV} & \boldmath$H_0^\dagger$ \\[1ex]
    \midrule
    GA & 10.88 & 1.49E-27 & 0.0100 & \xmark & 10.07 & 7.51E-24 & 0.0100 & \xmark\\[1ex]
    ACO & 9.20  & 3.47E-20 & 0.0125 & \xmark & 9.17  & 4.58E-20 & 0.0125 & \xmark \\[1ex]
    BPSO & 2.42  & 1.55E-02 & 0.0500 & \xmark & 5.32  & 1.04E-07 & 0.0250 & \xmark \\[1ex]
    CBPSO & 2.60  & 9.33E-03 & 0.0250 & \xmark & 4.60  & 4.19E-06 & 0.0500 & \xmark \\[1ex]
    chBPSO & 6.45  & 1.09E-10 & 0.0167 & \xmark & 6.33  & 2.38E-10 & 0.0167 & \xmark \\[1ex]
    % \midrule
    % \textbf{Outcome} & \multicolumn{3}{c}{\textit{All Hypotheses Rejected}} & \multicolumn{3}{c}{\textit{All Hypotheses Rejected}} \\
    
    \bottomrule
    \end{tabular}%
    \begin{tablenotes}
      \scriptsize
       \item $^\dagger$ $H_0$ denotes \textit{null-hypothesis}
    \end{tablenotes}
    \end{threeparttable}
   \end{adjustbox}
\end{table}%
%---------------------------------------------------------

First, the Friedman Two-way Analysis of Variance by Ranks~\cite{Sheskin:2003,Derrac:Salvador:2011} is applied to determine whether the performance of two or more compared algorithm is significantly different. For this purpose, the results obtained by the algorithms are ranked from `$1$' (\textit{best}) to `$6$' (\textit{worst}). Subsequently, the average value of ranks is determined over 40 independent runs. The test statistic and the corresponding \textit{p-value} are determined following the procedures outlined in~\cite{Sheskin:2003,Derrac:Salvador:2011} and are shown in Table~\ref{t:Fr1}. The \textit{p-values} obtained through the Friedman statistic strongly suggest a significant difference in the performance of the compared algorithms. Further, the average rankings obtained over 40 independent runs establish that 2D-UPSO is the best amongst the compared algorithms.

In the second step, a set of hypotheses is evaluated for multiple comparisons of 2D-UPSO with the other algorithms. Specifically, the evaluation of five interconnected null hypotheses is required to compare the 2D-UPSO with the five compared algorithms, \textit{i.e.}, GA, ACO, BPSO, CBPSO and chBPSO. Each null hypothesis ($H_0$) denotes that the algorithm being compared is significantly better than 2D-UPSO. The test statistic, \textit{$p-$value} and Adjusted p-values (APV) which are required to evaluate these interconnected hypotheses are determined following the procedure outlined in~\cite{Derrac:Salvador:2011,Garcia:Salvador:2010}. The Hommel's post-hoc procedure is employed to derive the APV from the \textit{$p-$value}. The outcome of the multiple comparisons for $95\%$ confidence interval are shown in Table~\ref{t:Fr2} (for k-NN) and Table~\ref{t:Fr3} (for NB). These results convincingly demonstrate that, among the compared algorithms, 2D-UPSO could obtain feature subsets with \textit{significantly} lower cardinality ($\xi$) and classification error, $J(\cdotp)$. 

%---------------------------------------------------------
\begin{table}[!t]
  \centering
  \small
  \caption{Outcome of the Hommel's Post-hoc Procedure for $95\%$ Confidence Interval (NB)}
  \label{t:Fr3}%
  \begin{adjustbox}{max width=0.8\textwidth}
  \begin{threeparttable}
    \begin{tabular}{ccccc|cccc}
    \toprule
    
    \multirow{2}{*}{\textbf{Algorithm}} & \multicolumn{4}{c}{\boldmath$J(\cdotp)$} & \multicolumn{4}{c}{\boldmath$\xi$} \\
    \cmidrule{2-9}  & \makecell{\textbf{Test} \\ \textbf{statistic}}     & \boldmath$p-$\textbf{value} & \textbf{APV} & \boldmath$H_0^\dagger$  & \makecell{\textbf{Test} \\ \textbf{statistic}}     & \boldmath$p-$\textbf{value} & \textbf{APV} & \boldmath$H_0^\dagger$ \\ [1ex]
    \midrule
    
    GA    & 10.79 & 3.97E-27 & 0.0100 & \xmark  & 8.46  & 2.76E-17 & 0.0100 & \xmark \\[1ex]
    ACO   & 9.71  & 2.70E-22 & 0.0125 & \xmark & 8.25  & 1.62E-16 & 0.0125 & \xmark \\[1ex]
    BPSO  & 2.69  & 7.16E-03 & 0.0500 & \xmark & 3.71  & 2.11E-04 & 0.0500 & \xmark \\[1ex]
    CBPSO & 3.08  & 2.09E-03 & 0.0250 & \xmark & 4.48  & 7.39E-06 & 0.0250 & \xmark\\[1ex]
    chBPSO & 6.54  & 5.99E-11 & 0.0167 & \xmark & 6.13  & 9.04E-10 & 0.0167 & \xmark \\[1ex]
    
    % \midrule
    % \textbf{Outcome} & \multicolumn{3}{c}{\textit{All Hypotheses Rejected}} & \multicolumn{3}{c}{\textit{All Hypotheses Rejected}} \\
    
    \bottomrule
    \end{tabular}%
    \begin{tablenotes}
      \scriptsize
       \item $^\dagger$ $H_0$ denotes \textit{null-hypothesis}
    \end{tablenotes}
    \end{threeparttable}
   \end{adjustbox}
\end{table}%
%---------------------------------------------------------
%---------------------------------------------------------
\begin{table}[!t]
  \centering
  \small
  \caption{Comparative analysis with SFFS}
  \label{t:sffs}%
  \begin{adjustbox}{max width=0.42\textwidth}
    \begin{tabular}{cccc}
    \toprule
    \multirow{2}[4]{*}{\textbf{IA}} & \multirow{2}[4]{*}{\boldmath$\xi_{best}$} & \multicolumn{2}{c}{\boldmath$J(\cdotp)$} \\
    \cmidrule{3-4}          &       & \textbf{SFFS}~\cite{Pudil:1994}  & \textbf{2D-UPSO} \\[0.8ex]
    \midrule
    \textit{k-NN}  & 10    & 0.0056 & \textbf{0.0029} \\[0.8ex]
    \textit{NB}    & 30    & 0.0070 & \textbf{0.0056} \\[0.8ex]
    \bottomrule
    \end{tabular}%
    \end{adjustbox}
\end{table}%
%---------------------------------------------------------
%---------------------------------------------------------
\subsubsection{Comparative analysis with SFFS}

The results of the comparative analysis with SFFS are shown in Table~\ref{t:sffs}. Note that SFFS requires \textit{a priori} specification of the subset cardinality, $\xi$. Since the cardinality of the optimum feature subset is not known, the cardinality of the best subset found by 2D-UPSO (out of $40$ runs) is determined and denoted as `$\xi_{best}$'. For example, the cardinality of the best subset found by 2D-UPSO with k-NN is $10$ and the same with NB is $30$. Hence, $\xi_{best}=10$ (with k-NN) and $\xi_{best}=30$ (with NB). Subsequently, SFFS is applied as a wrapper to both k-NN and NB to find the feature subset with the cardinality equal to `$\xi_{best}$'. Given the same cardinality $\xi_{best}$,the objective here is to determine whether the SFFS could identify the feature subset with the comparable or better accuracy than 2D-UPSO. As expected, the outcome of these tests (Table~\ref{t:sffs}) indicates that SFFS could not yield better feature subset with either k-NN or NB. 

%----------------------------------------------------------------------
\begin{table*}[!t]
    \centering
    \caption{Classification accuracy obtained with the reduced subsets (with k-NN)}
    \label{t:rknn}
    \begin{adjustbox}{max width=0.73\textwidth}
    \begin{threeparttable}
    \small
    \begin{tabular}{ccccccccc}
    \toprule
    \multicolumn{1}{c}{\textbf{\makecell{\textbf{Data Set} \\ \textbf{(SNR)}}}} & \textbf{\makecell{\textbf{All Features} \\ \boldmath $\Theta(U)$}} & \textbf{\textbf{Results$^\dagger$}} & \textbf{GA} & \textbf{ACO} & \textbf{BPSO} & \textbf{CBPSO} & \textbf{chBPSO} & \textbf{2D-UPSO} \\
    \midrule
    
    \multirow{2}{*}{$\infty \ dB$} & \multirow{2}{*}{$96.57$} & $\Theta(\cdotp)$ & 98.39 & 98.65 & 99.47 & 99.50 & 99.21 & \textbf{99.68} \\ [0.8ex]
          &       & $\theta(\cdotp)$ & 1.82  & 2.08  & 2.90  & 2.93  & 2.64  & \textbf{3.11} \\ [0.8ex]
    \midrule
    \multirow{2}{*}{$50 \ dB$} & \multirow{2}{*}{$94.82$} & $\Theta(\cdotp)$ & 96.78 & 97.51 & 97.83 & 97.51 & 97.54 & \textbf{98.86} \\ [0.8ex]
          &       & $\theta(\cdotp)$ & 1.96  & 2.69  & 3.02  & 2.69  & 2.72  & \textbf{4.04} \\ [0.8ex]
    \midrule
    \multirow{2}{*}{$45 \ dB$} & \multirow{2}{*}{$94.61$} & $\Theta(\cdotp)$ &  95.98 & 97.30 & 97.48 & 97.24 & 96.95 & \textbf{98.77} \\ [0.8ex]
          &       & $\theta(\cdotp)$ & 1.38  & 2.70  & 2.87  & 2.64  & 2.35  & \textbf{4.16} \\ [0.8ex]
    \midrule
    \multirow{2}{*}{$40 \ dB$} & \multirow{2}{*}{$93.55$} & $\Theta(\cdotp)$ & 95.40 & 96.31 & 96.86 & 96.45 & 95.95 & \textbf{98.48} \\ [0.8ex]
          &       & $\theta(\cdotp)$ & 1.85  & 2.76  & 3.31  & 2.91  & 2.41  & \textbf{4.93} \\ [0.8ex]
    \midrule
    \multirow{2}{*}{$35 \ dB$} & \multirow{2}{*}{$92.55$} & $\Theta(\cdotp)$ & 94.40 & 95.48 & 96.13 & 95.51 & 95.04 & \textbf{97.86} \\ [0.8ex]
          &       & $\theta(\cdotp)$ & 1.84  & 2.93  & 3.57  & 2.96  & 2.49  & \textbf{5.30} \\ [0.8ex]
    \midrule
    \multirow{2}{*}{$30 \ dB$} & \multirow{2}{*}{$91.86$} & $\Theta(\cdotp)$ & 93.70 & 94.34 & 94.58 & 94.28 & 93.79 & \textbf{96.34} \\ [0.8ex]
          &       & $\theta(\cdotp)$ & 1.85  & 2.49  & 2.72  & 2.43  & 1.93  & \textbf{4.48} \\ [0.8ex]
    \midrule
    \multirow{2}{*}{$25 \ dB$} & \multirow{2}{*}{$89.56$} & $\Theta(\cdotp)$ & 91.23 & 90.30 & 91.38 & 91.59 & 91.29 & \textbf{93.84} \\ [0.8ex]
          &       & $\theta(\cdotp)$ & 1.67  & 0.73  & 1.82  & 2.02  & 1.73  & \textbf{4.28} \\ [0.8ex]
    \midrule
    \multirow{2}{*}{$20 \ dB$} & \multirow{2}{*}{$85.46$} & $\Theta(\cdotp)$ & 86.04 & 86.13 & 86.66 & 86.45 & 86.45 & \textbf{88.48} \\ [0.8ex]
          &       & $\theta(\cdotp)$ & 0.59  & 0.67  & 1.20  & 0.99  & 1.00  & \textbf{3.02} \\ [0.8ex]
    \bottomrule
    \end{tabular}%
    \begin{tablenotes}
      \small
       \item $\dagger$ $\theta(\mathcal{X})=\Theta(\mathcal{X})-\Theta(U)$; the highest performance gain $\theta(\cdotp)$ is shown in bold-face
    \end{tablenotes}

    \end{threeparttable}
    \end{adjustbox}
%  \end{minipage}
% \end{table*}
% %----------------------------------------------------------------------
\medskip
% %----------------------------------------------------------------------
% \begin{table*}[!t]
    \centering
    \caption{Classification accuracy obtained with the reduced subsets (with NB)}
    \label{t:rnb}
    \begin{adjustbox}{max width=0.73\textwidth}
    \begin{threeparttable}
    \small
    \begin{tabular}{ccccccccc}
    \toprule
    \multicolumn{1}{c}{\textbf{\makecell{\textbf{Data Set} \\ \textbf{(SNR)}}}} & \textbf{\makecell{\textbf{All Features} \\ \boldmath $\Theta(U)$}} & \textbf{\textbf{Results$^\dagger$}} & \textbf{GA} & \textbf{ACO} & \textbf{BPSO} & \textbf{CBPSO} & \textbf{chBPSO} & \textbf{2D-UPSO} \\
    \midrule
    
    \multirow{2}{*}{$\infty \ dB$} & \multirow{2}{*}{98.01} & $\Theta(\cdotp)$ & 98.77 & 98.74 & 99.38 & 99.36 & 99.09 & \textbf{99.44} \\ [0.8ex]
          &       & $\theta(\cdotp)$ & 0.76  & 0.73  & 1.38  & 1.35  & 1.08  & \textbf{1.43} \\[0.8ex]
    \midrule
    \multirow{2}{*}{$50 \ dB$} & \multirow{2}{*}{97.30} & $\Theta(\cdotp)$ & 97.63 & 97.54 & 97.92 & 97.95 & 97.60 & \textbf{98.04} \\ [0.8ex]
          &       & $\theta(\cdotp)$ & 0.32  & 0.23  & 0.62  & 0.64  & 0.29  & \textbf{0.73} \\[0.8ex]
    \midrule
    \multirow{2}{*}{$45 \ dB$} & \multirow{2}{*}{97.04} & $\Theta(\cdotp)$ & 97.36 & 97.16 & 97.45 & 97.39 & 97.39 & \textbf{97.45} \\ [0.8ex]
          &       & $\theta(\cdotp)$ & 0.32  & 0.12  & 0.41  & 0.35  & 0.35  & \textbf{0.41} \\[0.8ex]
    \midrule
    \multirow{2}{*}{$40 \ dB$} & \multirow{2}{*}{96.34} & $\Theta(\cdotp)$ & 96.45 & 96.75 & 96.66 & 96.51 & 96.48 & \textbf{96.95} \\ [0.8ex]
          &       & $\theta(\cdotp)$ &  0.12  & 0.41  & 0.32  & 0.18  & 0.15  & \textbf{0.62} \\[0.8ex]
    \midrule
    \multirow{2}{*}{$35 \ dB$} & \multirow{2}{*}{95.66} & $\Theta(\cdotp)$ & 95.54 & 95.72 & 95.84 & 95.34 & 95.60 & \textbf{96.36} \\ [0.8ex]
          &       & $\theta(\cdotp)$ &  \underline{-0.12} & 0.06  & 0.18  & \underline{-0.32} & \underline{-0.06} & \textbf{0.70} \\[0.8ex]
    \midrule
    \multirow{2}{*}{$30 \ dB$} & \multirow{2}{*}{94.28} & $\Theta(\cdotp)$ & 94.78 & 94.05 & 94.58 & 93.72 & 93.99 & \textbf{95.10}\\ [0.8ex]
          &       & $\theta(\cdotp)$ & 0.50  & \underline{-0.24} & 0.29  & \underline{-0.56} & \underline{-0.30} & \textbf{0.82} \\[0.8ex]
    \midrule
    \multirow{2}{*}{$25 \ dB$} & \multirow{2}{*}{92.67} & $\Theta(\cdotp)$ & 92.58 & 92.29 & 92.52 & 92.11 & 92.35 & \textbf{93.87} \\ [0.8ex]
          &       & $\theta(\cdotp)$ & \underline{-0.09} & \underline{-0.38} & \underline{-0.15} & \underline{-0.56} & \underline{-0.32} & \textbf{1.20} \\[0.8ex]
    \midrule
    \multirow{2}{*}{$20 \ dB$} & \multirow{2}{*}{89.62} & $\Theta(\cdotp)$ & 89.94 & 89.68 & 90.15 & 89.77 & 90.00 & \textbf{91.41} \\ [0.8ex]
          &       & $\theta(\cdotp)$ & 0.32  & 0.06  & 0.53  & 0.14  & 0.38  & \textbf{1.79} \\[0.8ex]
    \bottomrule
    \end{tabular}%
    
    \begin{tablenotes}
      \small
       \item $\dagger$ $\theta(\mathcal{X})=\Theta(\mathcal{X})-\Theta(U)$; the highest performance gain $\theta(\cdotp)$ is shown in bold-face
    \end{tablenotes}
    
    \end{threeparttable}
    \end{adjustbox}
\end{table*}
%----------------------------------------------------------------------

%---------------------------------------------------------
\subsection{Stage-II : Robustness of the Reduced Subsets}
\label{s:test}

To evaluate the robustness of the reduced subsets against measurement noise, seven different levels of noise with SNR$=[50, 45, 40, 35, 30, 25, 20] dB$ have been added to the PQ events. Given that the noise introduced by the measurement chains cannot be \textit{a priori} estimated, the objective here is to compare the classification performance of the reduced subsets and the original feature set ($U$) at various noise levels. 

Note that the feature selection has been carried out using only \textit{pure} PQ events, \textit{i.e.} with SNR=$\infty \ dB$. Further, each run of the compared algorithms provides a different feature subset due to the inherent stochastic nature. To ensure the fair comparison, for each algorithm the feature subset with the minimum classification error, $J(\cdotp)$, out of $40$ runs is selected.

For each feature subset under consideration the average classification accuracy after $10$-fold stratified cross-validation, `$\Theta(\cdotp)$', is recorded at each noise level. This is given by,
%---------------------------------------------------------
\begin{linenomath*}
\begin{align}
    \label{eq:theta}
        \Theta(\mathcal{X}) & = \frac{1}{10} \times \sum \limits_{f=1}^{10} \eta_f, \\
        \text{where, } \eta_f & = \frac{\textit{correctly classified events in the $f^{th}$-fold}}{\textit{total no. of events in the $f^{th}$-fold}} \nonumber
\end{align}
\end{linenomath*}
%---------------------------------------------------------
`$\mathcal{X}$' and `$\Theta(\mathcal{X})$' respectively denote the reduced subset under consideration and the corresponding classification accuracy. Note that, $\Theta(\mathcal{X})=1-J(\mathcal{X})$. 

In order to evaluate the \textit{`robustness'} of a given feature subset, $\mathcal{X}$, its classification accuracy, $\Theta(\mathcal{X})$, is compared with that of the original full feature set, $\Theta(U)$. As the feature selection was carried out using the `pure' dataset (\textit{i.e.}, SNR$=\infty \ dB$), the performance of all reduced subsets is better compared to $U$ at this noise level. The objective here is to investigate whether the reduced subsets could maintain improved performance in the presence of various other levels of noise. Essentially, at each noise level, we are interested in finding out whether $\theta(\mathcal{X}) \geq 0$, which is given by,
%---------------------------------------------------------
\begin{linenomath*}
\begin{align}
\label{eq:theta2}
    \theta(\mathcal{X})=\Theta(\mathcal{X}) - \Theta(U)
\end{align}
\end{linenomath*}
%---------------------------------------------------------

The classification accuracy, $\Theta(\cdotp)$ and the performance difference $\theta(\cdotp)$ of the reduced subsets are shown in Table~\ref{t:rknn} (with k-NN) and Table~\ref{t:rnb} (with NB). The metric $\theta(\cdotp)$ in (\ref{eq:theta2}) indicates the degree of improvement over $\Theta(U)$, \textit{i.e.}, a higher positive value of $\theta(\cdotp)$ is desirable. It is clear that the reduced subset obtained by 2D-UPSO could achieve the highest $\theta(\cdotp)$ with both k-NN (Table~\ref{t:rknn}) and NB (Table~\ref{t:rnb}). By integrating the cardinality information into the search process, 2D-UPSO could find robust and effective feature subsets. These results convincingly demonstrate that with a proper feature selection approach, it is possible to obtain a robust feature subset that can yield enhanced performance even in the presence of various levels of measurement noise.

Further, the results in Tables~\ref{t:rknn} and~\ref{t:rnb} clearly show the influence of the induction algorithms on the search landscape. For example, with k-NN, all algorithms could identify robust feature subset, \textit{i.e.}, \textit{a positive $\theta(\cdotp)$ at all noise levels}, as seen in Table~\ref{t:rknn}. In contrast, with NB, only 2D-UPSO could yield robust feature subset (Table~\ref{t:rnb}). These results further underline the importance of a feature selection approach.

%----------------------------------------------------------------------
%-------- CONTRAST ESTIMATION (k-NN)
%----------------------------------------------------------------------
\begin{table}[!t]
\centering
\small
\caption{Contrast Estimation(k-NN)}
\label{t:ceknn}
\begin{adjustbox}{max width=0.65\textwidth}
\begin{tabular}{ccccccc}
\toprule
 &\textbf{GA} & \textbf{ACO} & \textbf{BPSO} & \textbf{CBPSO} & \textbf{chBPSO} & \textbf{2D-UPSO}\\
\midrule
\textbf{GA}     & 0 & -0.6658 &-1.078 & -0.8050 & -0.5517 & -2.759\\[0.5ex]
\textbf{ACO}    & \textbf{0.6658} & 0 & -0.4125 & -0.1392 & \textbf{0.1142} & -2.093\\[0.5ex]
\textbf{BPSO}   & \textbf{1.078} & \textbf{0.4125} & 0 & \textbf{0.2733} & \textbf{0.5267} & -1.681\\[0.5ex]
\textbf{CBPSO}  & \textbf{0.8050} & \textbf{0.1392} & -0.2733 & 0 & \textbf{0.2533} & -1.954\\[0.5ex]
\textbf{chBPSO} & \textbf{0.5517} & -0.1142 & -0.5267 & -0.2533 & 0 & -2.208\\[0.5ex]
\textbf{2D-UPSO}& \textbf{2.759} & \textbf{2.093} & \textbf{1.681} & \textbf{1.954} & \textbf{2.208} & 0 \\[0.5ex]
\bottomrule
\end{tabular}
\end{adjustbox}
% \end{table}
%----------------------------------------------------------------------
\medskip
\bigskip
%----------------------------------------------------------------------
%-------- CONTRAST ESTIMATION (NB)
%----------------------------------------------------------------------
% \begin{table}[!htp]
\centering
\small
\caption{Contrast Estimation(NB)}
\label{t:cenb}
\begin{adjustbox}{max width=0.65\textwidth}
\begin{tabular}{ccccccc}
\toprule
 &\textbf{GA} & \textbf{ACO} & \textbf{BPSO} & \textbf{CBPSO} & \textbf{chBPSO} & \textbf{2D-UPSO}\\
\midrule

\textbf{GA}         & 0 & \textbf{0.0925} & -0.2042 & \textbf{0.1025} & \textbf{0.0017} & -0.6025\\[0.5ex]
\textbf{ACO}        & -0.0925 & 0 & -0.2967 & \textbf{0.01} & -0.0908 & -0.6950\\[0.5ex]
\textbf{BPSO}       & \textbf{0.2042} & \textbf{0.2967} & 0 & \textbf{0.3067} & \textbf{0.2058} &-0.3983\\[0.5ex]
\textbf{CBPSO}      & -0.1025 & -0.01 & -0.3067 & 0 & -0.1008 & -0.7050\\[0.5ex]
\textbf{chBPSO}     & -0.0017 & \textbf{0.0908} & -0.2058 & \textbf{0.1008} & 0 & -0.6042\\[0.5ex]
\textbf{2D-UPSO}    & \textbf{0.6025} & \textbf{0.6950} & \textbf{0.3983} & \textbf{0.7050} & \textbf{0.6042} & 0 \\[0.5ex]
\bottomrule
\end{tabular}
\end{adjustbox}
\end{table}
%----------------------------------------------------------------------

Finally, the statistical significance of the results shown in Table~\ref{t:rknn} and Table~\ref{t:rnb} is determined. For this purpose, it is not feasible to apply the multiple non-parametric statistical comparisons (similar to Section~\ref{s:Fried}), since the number of datasets is relatively small~\cite{Derrac:Salvador:2011}. Therefore, Contrast Estimation based on medians~\cite{Garcia:Salvador:2010,Derrac:Salvador:2011} is applied to compare the algorithms. This test essentially estimates a quantitative performance difference over multiple datasets for all possible pairs of algorithms. 

The outcomes of this test are shown in Table~\ref{t:ceknn} (with k-NN) and Table~\ref{t:cenb} (with NB). Note that a higher positive value of the estimator is desirable for this test. For instance, with k-NN, the contrast estimator for ACO is positive with respect to GA ($0.6658$, Table~\ref{t:ceknn}) and chBPSO ($0.1142$, Table~\ref{t:ceknn}) which indicate that the ACO could yield better subset in comparison to GA and chBPSO. For each of the algorithms, the positive outcomes are shown in boldface in Table~\ref{t:ceknn} and~\ref{t:cenb}. As seen in Table~\ref{t:ceknn} and~\ref{t:cenb}, the estimator for 2D-UPSO is positive for each pairwise comparison, which further highlights enhanced search performance of 2D-UPSO. Furthermore, the shift in the search landscape can indirectly be illustrated by the search behavior of GA; For instance, with k-NN, the estimator values for GA are negative with respect to all the algorithms (Table~\ref{t:ceknn}). In contrast, with NB, the positive estimators for GA are obtained with respect to three algorithms (ACO, CBPSO and chBPSO, Table~\ref{t:cenb}).

%%%-------------------------- New Section ---------------------------------
\section{Discussion}
\label{s:discuss}

The following observations are inferred from the results of the comparative evaluation (Section~\ref{s:ce}) and the robustness evaluation (Section~\ref{s:test}):
\medskip
\begin{itemize}%[leftmargin=5mm,labelindent=5mm,labelsep=2mm,itemsep=0.8\baselineskip]
    \item The feature selection wrappers are often criticized for computational complexity. However, as the results of this investigation suggest, wrappers can yield a significant reduction in the feature subset size while improving the classification performance. For instance, the compared algorithms could reduce the original feature set in the range of $52\%-79\%$ (with k-NN, Table~\ref{t:res:knn}) and $50\%-63\%$ (with NB, Table~\ref{t:res:nb}) while improving the classification performance in the range of $53\%-87\%$ (with k-NN, Table~\ref{t:res:knn}) and $32\%-69\%$ (with NB, Table~\ref{t:res:nb}). Since the feature selection is performed only once, the wrapper approach is highly recommended for the PQ events. 
    \item With an effective search strategy, it is possible to identify a feature subset which is \textit{robust} against the measurement noise. In this study, the worst case scenario from the perspective of PQ events has been simulated, \textit{i.e.}, the feature selection is carried out using only \textit{`pure'} PQ events (\textit{i.e.}, SNR=$\infty$ dB) and subsequently the reduced subsets are evaluated under various levels of measurement noise. Nevertheless, under this scenario, 2D-UPSO could identify robust feature subsets with both k-NN and NB. 
    \item The results give empirical evidence for the hypothesis that the nature of the induction algorithm does affect the feature selection landscape. For example, all of the compared algorithms could yield robust feature subset with k-NN; however, only 2D-UPSO could identify a robust feature subset with NB. This further underlines the need for a \textit{wrapper} based feature selection approach.
\end{itemize}

%%%-------------------------- New Section ---------------------------------
\section{Conclusions}
\label{s:con}

In this study, the issue of feature selection has comprehensively been investigated in the context of PQ event identification. In particular, the search performance of the Two-dimensional learning approach (2D-UPSO) and six other feature selection wrappers has been compared considering fourteen distinct classes of PQ events. Further, the \textit{robustness} of the reduced feature subsets has been defined and evaluated under seven different levels of measurement noise. The results of the comparative evaluation convincingly demonstrate that 2D-UPSO can identify significantly better and \textit{robust} feature subsets for PQ events. The key distinctive property of 2D learning is the integration of information about the subset size into the learning framework. This has been shown to lead to a significant improvement in the search performance in comparison to the other well-known algorithms, \textit{e.g.}, GA, ACO and BPSO.

Without loss of generality, this investigation is based on the assumption that the induction algorithm for the PQ event identification is pre-fixed. If this assumption does not hold or a \textit{generalized} reduced feature subset is desired, then the \textit{filter} based feature selection approaches are more appropriate. Hence, a detailed comparative investigation of the different filter approaches such as Mutual Information, Minimum Redundancy Maximum Relevance, Correlation based feature selection may prove to be very useful to both the practicing engineers and the PQ researchers. This could be the subject of further research.

%%%-------------------------- New Section ---------------------------------
\linespread{1}
\section*{Acknowledgement}
Faizal Hafiz is thankful to Education New Zealand for supporting this research through the New Zealand International Doctoral Research Scholarship (NZIDRS).
%%%-------------------------- Appendix ---------------------------------
\appendix
\section{Illustrative Example of 2D Learning}
\label{s:appexample}
  
Consider a dataset having $5$ number of features ($n=5$). For this dataset, let the position of the $i^{th}$ particle `$\beta_i$' and the learning exemplar `$\alpha$' at particular search iteration `$t$' be given by,
%---------------------------------------------------------
\begin{align}
\label{eq:exapos}
\beta_i & =\begin{bmatrix} 1 & 1 & 0 & 1 & 0 \end{bmatrix}, \ \ \alpha = \begin{bmatrix} 1 & 1 & 1 & 0 & 1 \end{bmatrix}
\end{align}
%---------------------------------------------------------

\subsection*{Evaluation of the Learning Sets}

The learning sets derived from $\alpha$ and $\beta_i$, as per the Algorithm~\ref{alg:learningset}, are as follows: 
\begin{enumerate}
    \item Set cardinality learning sets to null-vector:\\
    $\varphi_{\alpha} = \varphi_i = \begin{bmatrix} 0 & 0 & 0 & 0 & 0 \end{bmatrix}$
    \smallskip
    \item Cardinality of $\alpha$ and $\beta_i$:\\
    $\xi_{\alpha} = \sum \limits_{m=1}^{n} \alpha_m = 4$ and $\xi_i= \sum \limits_{m=1}^{n} \beta_{i,m} = 3$.
    \smallskip
    \item Set the $\xi^{th}$ bit of $\varphi$ to `$1$':\\
    $\varphi_{\alpha} = \begin{bmatrix} 0 & 0 & 0 & 1 & 0 \end{bmatrix}$ and $\varphi_i = \begin{bmatrix} 0 & 0 & 1 & 0 & 0 \end{bmatrix}$
    \smallskip
    \item Evaluate feature learning sets:\\
    $\psi_{\alpha} = \{ \alpha \wedge \overline{\beta}_i \} = \begin{bmatrix} 0 & 0 & 1 & 0 & 1 \end{bmatrix}$ and $\psi_i = \beta_i = \begin{bmatrix} 1 & 1 & 0 & 1 & 0 \end{bmatrix}$
    \smallskip
    \item Evaluate the final learning sets:\\
        $\mathcal{L}_{\alpha} = \begin{bmatrix} \varphi_{\alpha} \\ \psi_{\alpha} \end{bmatrix} = \begin{bmatrix} 0 & 0 & 0 & 1 & 0 \\ 0 & 0 & 1 & 0 & 1 \end{bmatrix}$ and $\mathcal{L}_{i} = \begin{bmatrix} \varphi_{i} \\ \psi_{i} \end{bmatrix} = \begin{bmatrix} 0 & 0 & 1 & 0 & 0 \\ 1 & 1 & 0 & 1 & 0 \end{bmatrix}$ 
\end{enumerate}

\subsection*{Position Update}

To understand the position update procedure, assume that the velocity of the $i^{th}$ particle is given by,
%---------------------------------------------------------
\begin{align}
\label{eq:exavel}
\mathcal{V}_i = \begin{bmatrix} 0.82 & 2.53 & 2.22 & 0.28 & 0.95 \\ 1.61 & 1.88 & 0.80 & 1.33 & 2.88 \end{bmatrix}
\end{align}
%---------------------------------------------------------
The new position of the $i^{th}$ particle is determined as per the procedure outlined in Algorithm~\ref{fig:posprop}:
\begin{enumerate}
    \item Set the new position to a \textit{null}-vector:
    $\beta_i=\begin{bmatrix} 0 & 0 & 0 & 0 & 0 \end{bmatrix}$
    \item Isolate the \textit{selection likelihoods} of cardinality and features, $\mathcal{V}_i = \begin{bmatrix} \rho \\ \sigma \end{bmatrix}$:\\
    $\rho = \begin{bmatrix} 0.82 & 2.53 & 2.22 & 0.28 & 0.95 \end{bmatrix}$ and $\sigma = \begin{bmatrix} 1.61 & 1.88 & 0.80 & 1.33 & 2.88 \end{bmatrix}$
    \smallskip
    \item Evaluate the accumulative likelihoods:\\
    $\rho_\Sigma= \begin{bmatrix} 0.82 & 3.35 & 5.57 & 5.85 & 6.80 \end{bmatrix}$ 
    \smallskip
    \item Let \textit{random number}, $r \in [0,\rho_{\Sigma,n}]$, be `$3.1$'
    \smallskip
    \item Set $\xi=j$ such that $\rho_{\Sigma,j-1}<r<\rho_{\Sigma,j}$:\\
    $\xi_i = 2$ as $\rho_{\Sigma,1}<r<\rho_{\Sigma,2}$ 
    \smallskip
    \item Rank the features on the basis of $\sigma$:\\
            $\sigma = \begin{bmatrix} 1.61 & 1.88 & 0.80 & 1.33 & 2.88 \end{bmatrix}$ which gives, $\tau = \begin{bmatrix} 3 & 2 & 5 & 4 & 1 \end{bmatrix}$
    \smallskip
    \item Select features with a rank less than or equal to $\xi_i$:\\
            $\xi_i=2$ and $\tau = \begin{bmatrix} 3 & 2 & 5 & 4 & 1 \end{bmatrix}$ which gives, $\beta_i= \begin{bmatrix} 0 & 1 & 0 & 0 & 1 \end{bmatrix}$
\end{enumerate}

\section{ Determination of DWT Decomposition Levels}
\label{s:decomplvl}
The decomposition level for DWT can be determined from the following thumb-rule:%$\frac{F_s}{2^{L+1}} \leq f \leq \frac{F_s}{2^L}$. \\
\begin{align}
    \frac{F_s}{2^{D+1}} \leq f \leq \frac{F_s}{2^D} \nonumber
\end{align}
where, `$F_s$' is sampling frequency, `$f$' is the fundamental frequency and `$D$' denotes the required decomposition levels. 

\section{Statistical Functions used for Feature Extraction}
\label{s:app}
\begin{center}
{
$f_1$: $ \min\limits_{i=1 \dots N_l} c_i^k$, $f_2$: $ \max\limits_{i=1 \dots N_l} c_i^k$, $f_3$: Median; $f_4$: $ \sum\limits_{i=1}^{N_l} |c_i^k|^2$, $f_5$: $\frac{1}{N_l} \sum \limits_{i=1}^{N_l}c_i^k $ , $f_{6}$: $-\sum \limits_{i=1}^{N_l} p_i \log p_i$, $p_i = \frac{|c_i^k|^{2}}{f_4}$\\ [1.2ex]
$f_7$: $\frac{\sum \limits_{i=1}^{N_l} ({c_i^k- f_5})^3}{(N_l-1)\;\sigma^3}$, $f_8$: $\sigma = \sqrt{\frac{\sum \limits_{i=1}^{N_l} (c_i^k- f_5)^2}{(N_l-1)}} $, $f_9$: $ \frac{1}{N_l} \sum \limits_{i=1}^{N_l} |c_i^k - f_5|$, $f_{10}$: $ \frac{\sum \limits_{i=1}^{N_l} (c_i^k- f_5)^2}{(N_l-1)}$, 
$f_{11}$: $\frac{1}{(N_l-1) \; \sigma^4} \sum \limits_{i=1}^{N_l} ({c_i^k - f_5})^4 $
}
\end{center}
 
 where, $C^j=\{ c_1^j,c_2^j, \dots c_{N_l}^j \}$ and `$N_l$' denotes number of \textit{detail/approximation} coefficients at the $j^{th}$ decomposition level.

\bibliographystyle{elsarticle-num}

\newcommand{\noop}[1]{}
\begin{thebibliography}{10}
\expandafter\ifx\csname url\endcsname\relax
  \def\url#1{\texttt{#1}}\fi
\expandafter\ifx\csname urlprefix\endcsname\relax\def\urlprefix{URL }\fi
\expandafter\ifx\csname href\endcsname\relax
  \def\href#1#2{#2} \def\path#1{#1}\fi

\bibitem{IEEE:1159}
IEEE, {IEEE} recommended practice for monitoring electric power quality, IEEE
  Std 1159-2009 (Revision of IEEE Std 1159-1995) (2009) c1--81.

\bibitem{Santoso:1996}
S.~Santoso, E.~J. Powers, W.~M. Grady, P.~Hofmann, Power quality assessment via
  wavelet transform analysis, IEEE Transactions on Power Delivery 11~(2) (1996)
  924--930.

\bibitem{Santoso:2000}
S.~Santoso, W.~M. Grady, E.~J. Powers, J.~Lamoree, S.~C. Bhatt,
  Characterization of distribution power quality events with fourier and
  wavelet transforms, IEEE Transactions on Power Delivery 15~(1) (2000)
  247--254.

\bibitem{Angrisani:1998}
L.~Angrisani, P.~Daponte, M.~D'apuzzo, A.~Testa, A measurement method based on
  the wavelet transform for power quality analysis, IEEE Transactions on Power
  Delivery 13~(4) (1998) 990--998.

\bibitem{Gaouda:Salama:1999}
A.~M. Gaouda, M.~M.~A. Salama, M.~R. Sultan, A.~Y. Chikhani, Power quality
  detection and classification using wavelet-multiresolution signal
  decomposition, IEEE Transactions on Power Delivery 14~(4) (1999) 1469--1476.

\bibitem{Mahela:Shaik:2015}
O.~P. Mahela, A.~G. Shaik, N.~Gupta, A critical review of detection and
  classification of power quality events, Renewable and Sustainable Energy
  Reviews 41 (2015) 495--505.

\bibitem{Khokhar:Zin:2015}
S.~Khokhar, A.~A. B.~M. Zin, A.~S.~B. Mokhtar, M.~Pesaran, A comprehensive
  overview on signal processing and artificial intelligence techniques
  applications in classification of power quality disturbances, Renewable and
  Sustainable Energy Reviews 51 (2015) 1650--1663.

\bibitem{Mahela:Shaik:2017}
O.~P. Mahela, A.~G. Shaik, Recognition of power quality disturbances using
  {S}-transform based ruled decision tree and fuzzy c-means clustering
  classifiers, Applied Soft Computing 59 (2017) 243 -- 257.

\bibitem{Kumar:Singh:2015}
R.~Kumar, B.~Singh, D.~Shahani, A.~Chandra, K.~Al-Haddad, Recognition of
  power-quality disturbances using {S}-transform-based {ANN} classifier and
  rule-based decision tree, IEEE Transactions on Industry Applications 51~(2)
  (2015) 1249--1258.

\bibitem{Liu:Cui:2015}
Z.~Liu, Y.~Cui, W.~Li, A classification method for complex power quality
  disturbances using {EEMD} and rank wavelet {SVM}, IEEE Transactions on Smart
  Grid 6~(4) (2015) 1678--1685.

\bibitem{Biswal:Dash:2013}
M.~Biswal, P.~K. Dash, Detection and characterization of multiple power quality
  disturbances with a fast {S}-transform and decision tree based classifier,
  Digital Signal Processing 23~(4) (2013) 1071--1083.

\bibitem{Naik:Hafiz:2016}
C.~Naik, F.~Hafiz, A.~Swain, A.~Kar, Classification of power quality events
  using wavelet packet transform and extreme learning machine, in: Power
  Electronics Conference (SPEC), IEEE Annual Southern, IEEE, 2016, pp. 1--6.

\bibitem{Marill:Green:1963}
T.~Marill, D.~Green, On the effectiveness of receptors in recognition systems,
  IEEE Transactions on Information Theory 9~(1) (1963) 11--17.

\bibitem{Blum:Langley:1997}
A.~L. Blum, P.~Langley, Selection of relevant features and examples in machine
  learning, Artificial Intelligence 97~(1) (1997) 245--271.

\bibitem{Dash:Liu:1997}
M.~Dash, H.~Liu, Feature selection for classification, Intelligent Data
  Analysis 1~(3) (1997) 131--156.

\bibitem{Guyon:Isabelle:2003}
I.~Guyon, A.~Elisseeff, An introduction to variable and feature selection,
  Journal of Machine Learning Research 3~(Mar) (2003) 1157--1182.

\bibitem{Xue:Zhang:2016}
B.~Xue, M.~Zhang, W.~N. Browne, X.~Yao, A survey on evolutionary computation
  approaches to feature selection, IEEE Transactions on Evolutionary
  Computation 20~(4) (2016) 606--626.

\bibitem{Panigrahi:2009}
B.~Panigrahi, V.~R. Pandi, Optimal feature selection for classification of
  power quality disturbances using wavelet packet-based fuzzy k-nearest
  neighbour algorithm, IET Generation, Transmission \& Distribution 3~(3)
  (2009) 296--306.

\bibitem{Gunal:2009}
S.~Gunal, O.~N. Gerek, D.~G. Ece, R.~Edizkan, The search for optimal feature
  set in power quality event classification, Expert Systems with Applications
  36~(7) (2009) 10266--10273.

\bibitem{Lee:Shen:2011}
C.~Y. Lee, Y.~X. Shen, Optimal feature selection for power-quality disturbances
  classification, IEEE Transactions on Power Delivery 26~(4) (2011) 2342--2351.

\bibitem{Manimala:Selvi:2011}
K.~Manimala, K.~Selvi, R.~Ahila, Hybrid soft computing techniques for feature
  selection and parameter optimization in power quality data mining, Applied
  Soft Computing 11~(8) (2011) 5485 -- 5497.

\bibitem{Manimala:Selvi:2012}
K.~Manimala, K.~Selvi, R.~Ahila, Optimization techniques for improving power
  quality data mining using wavelet packet based support vector machine,
  Neurocomputing 77~(1) (2012) 36--47.

\bibitem{Ericsti:2013}
H.~Eri{\c{s}}ti, {\"O}.~Y{\i}ld{\i}r{\i}m, B.~Eri{\c{s}}ti, Y.~Demir, Optimal
  feature selection for classification of the power quality events using
  wavelet transform and least squares support vector machines, International
  Journal of Electrical Power \& Energy Systems 49 (2013) 95--103.

\bibitem{Dalai:Chatterjee:2013}
S.~Dalai, B.~Chatterjee, D.~Dey, S.~Chakravorti, K.~Bhattacharya,
  Rough-set-based feature selection and classification for power quality
  sensing device employing correlation techniques, IEEE Sensors Journal 13~(2)
  (2013) 563--573.

\bibitem{Hajian:Foroud:2014a}
M.~Hajian, A.~A. Foroud, A.~A. Abdoos, New automated power quality recognition
  system for online/offline monitoring, Neurocomputing 128 (2014) 389--406.

\bibitem{Hajian:Foroud:2014b}
M.~Hajian, A.~A. Foroud, A new hybrid pattern recognition scheme for automatic
  discrimination of power quality disturbances, Measurement 51 (2014) 265--280.

\bibitem{Abdoos:Mianaei:2016}
A.~A. Abdoos, P.~K. Mianaei, M.~R. Ghadikolaei, Combined {VMD-SVM} based
  feature selection method for classification of power quality events, Applied
  Soft Computing 38 (2016) 637 -- 646.

\bibitem{Hafiz:Swain:2017}
F.~Hafiz, A.~Swain, C.~Naik, N.~Patel, Feature selection for power quality
  event identification, in: TENCON 2017 - 2017 IEEE Region 10 Conference, 2017,
  pp. 2984--2989.

\bibitem{Khokhar:Zin:2017}
S.~Khokhar, A.~A.~M. Zin, A.~P. Memon, A.~S. Mokhtar, A new optimal feature
  selection algorithm for classification of power quality disturbances using
  discrete wavelet transform and probabilistic neural network, Measurement 95
  (2017) 246--259.

\bibitem{Singh:Singh:2017}
U.~Singh, S.~N. Singh, Optimal feature selection via {NSGA-II} for power
  quality disturbances classification, IEEE Transactions on Industrial
  Informatics 14~(7) (2018) 2994--3002.

\bibitem{Siedlecki:Sklansky:1989}
W.~Siedlecki, J.~Sklansky, A note on genetic algorithms for large-scale feature
  selection, Pattern Recognition Letters 10~(5) (1989) 335--347.

\bibitem{Kohavi:1994}
R.~Kohavi, Feature subset selection as search with probabilistic estimates, in:
  AAAI fall symposium on relevance, Vol. 224, 1994, pp. 109--113.

\bibitem{Kohavi:John:1997}
R.~Kohavi, G.~H. John, Wrappers for feature subset selection, Artificial
  intelligence 97~(1) (1997) 273--324.

\bibitem{Yang:Honavar:1998}
J.~Yang, V.~Honavar, Feature subset selection using a genetic algorithm, IEEE
  Intelligent Systems and their Applications 13~(2) (1998) 44--49.

\bibitem{Hafiz:Swain:2017a}
F.~Hafiz, A.~Swain, N.~Patel, C.~Naik, A two-dimensional ({2-D}) learning
  framework for particle swarm based feature selection, Pattern Recognition 76
  (2018) 416 -- 433.

\bibitem{Biswal:Dash:2013a}
M.~Biswal, P.~K. Dash, Measurement and classification of simultaneous power
  signal patterns with an {S}-transform variant and fuzzy decision tree, IEEE
  Transactions on Industrial Informatics 9~(4) (2013) 1819--1827.

\bibitem{Cover:Van:1977}
T.~M. Cover, J.~M. Van~Campenhout, On the possible orderings in the measurement
  selection problem, IEEE Transactions on Systems, Man, and Cybernetics 7~(9)
  (1977) 657--661.

\bibitem{Liu:Yu:2005}
H.~Liu, L.~Yu, Toward integrating feature selection algorithms for
  classification and clustering, IEEE Transactions on knowledge and data
  engineering 17~(4) (2005) 491--502.

\bibitem{Shang:Wang:2016}
R.~Shang, W.~Wang, R.~Stolkin, L.~Jiao, Subspace learning-based graph
  regularized feature selection, Knowledge-Based Systems 112 (2016) 152--165.

\bibitem{Shang:Wang:2018}
R.~Shang, W.~Wang, R.~Stolkin, L.~Jiao, Non-negative spectral learning and
  sparse regression-based dual-graph regularized feature selection, IEEE
  transactions on cybernetics 48~(2) (2018) 793--806.

\bibitem{Whitney:1971}
A.~W. Whitney, A direct method of nonparametric measurement selection, IEEE
  Transactions on Computers 100~(9) (1971) 1100--1103.

\bibitem{Pudil:1994}
P.~Pudil, J.~Novovi{\v{c}}ov{\'a}, J.~Kittler, Floating search methods in
  feature selection, Pattern Recognition Letters 15~(11) (1994) 1119--1125.

\bibitem{Somol:Pudil:1999}
P.~Somol, P.~Pudil, J.~Novovi{\v{c}}ov{\'a}, P.~Pacl{\i}k, Adaptive floating
  search methods in feature selection, Pattern Recognition Letters 20~(11)
  (1999) 1157--1163.

\bibitem{Narendra:1977}
P.~M. Narendra, K.~Fukunaga, A branch and bound algorithm for feature subset
  selection, IEEE Transactions on Computers 100~(9) (1977) 917--922.

\bibitem{Yu:Yuan:1993}
B.~Yu, B.~Yuan, A more efficient branch and bound algorithm for feature
  selection, Pattern Recognition 26~(6) (1993) 883--889.

\bibitem{Hafiz:Abdennour:2013}
F.~M. Hafiz, A.~Abdennour, A team-oriented approach to particle swarms, Applied
  Soft Computing 13~(9) (2013) 3776 -- 3791.

\bibitem{Hafiz:Abdennour:2016}
F.~Hafiz, A.~Abdennour, Particle swarm algorithm variants for the quadratic
  assignment problems-a probabilistic learning approach, Expert Systems with
  Applications 44 (2016) 413--431.

\bibitem{UPSO1}
K.~E. Parsopoulos, M.~N. Vrahatis, {UPSO}: A unified particle swarm
  optimization scheme, Lecture Series on Computer and Computational Sciences
  1~(5) (2004) 868--873.

\bibitem{Shi:Eberhart:1998}
Y.~Shi, R.~Eberhart, A modified particle swarm optimizer, in: IEEE
  International Conference on Evolutionary Computation, IEEE, 1998, pp. 69--73.

\bibitem{Kennedy:Mendes:2002}
J.~Kennedy, R.~Mendes, Population structure and particle swarm performance, in:
  Proceedings of the Congress on Evolutionary Computation (CEC '02), Vol.~2,
  2002, pp. 1671--1676.

\bibitem{Hafiz:Swain:2019}
F.~Hafiz, A.~Swain, C.~Naik, S.~Abecrombie, A.~Eaton, Identification of power
  quality events: selection of optimum base wavelet and machine learning
  algorithm, IET Science, Measurement \& Technology 13 (2019) 260--271.

\bibitem{Witten:Frank:Hall:2016}
I.~H. Witten, E.~Frank, M.~A. Hall, C.~J. Pal, Data Mining: Practical machine
  learning tools and techniques, Morgan Kaufmann, 2016.

\bibitem{Kudo:Sklansky:2000}
M.~Kudo, J.~Sklansky, Comparison of algorithms that select features for pattern
  classifiers, Pattern Recognition 33~(1) (2000) 25--41.

\bibitem{Yu:Gu:2009}
H.~Yu, G.~Gu, H.~Liu, J.~Shen, J.~Zhao, A modified ant colony optimization
  algorithm for tumor marker gene selection, Genomics, Proteomics \&
  Bioinformatics 7~(4) (2009) 200--208.

\bibitem{Chen:Chen:2013}
B.~Chen, L.~Chen, Y.~Chen, Efficient ant colony optimization for image feature
  selection, Signal Processing 93~(6) (2013) 1566--1576.

\bibitem{Kennedy:Eberhart:1997}
J.~Kennedy, R.~C. Eberhart, A discrete binary version of the particle swarm
  algorithm, in: Proccedings of IEEE International Conference on Computational
  Cybernetics and Simulation, Vol.~5, 1997, pp. 4104--4108.

\bibitem{Chuang:Tsai:2011}
L.~Y. Chuang, S.~W. Tsai, C.~H. Yang, Improved binary particle swarm
  optimization using catfish effect for feature selection, Expert Systems with
  Applications 38~(10) (2011) 12699--12707.

\bibitem{Chuang:Yang:2011}
L.~Y. Chuang, C.~H. Yang, J.~C. Li, Chaotic maps based on binary particle swarm
  optimization for feature selection, Applied Soft Computing 11~(1) (2011)
  239--248.

\bibitem{Derrac:Salvador:2011}
J.~Derrac, S.~Garc{\'\i}a, D.~Molina, F.~Herrera, A practical tutorial on the
  use of nonparametric statistical tests as a methodology for comparing
  evolutionary and swarm intelligence algorithms, Swarm and Evolutionary
  Computation 1~(1) (2011) 3--18.

\bibitem{Sheskin:2003}
D.~Sheskin, Handbook of parametric and nonparametric statistical procedures,
  3rd Edition, Chapman \& Hall/CRC, Boca Raton, FL, 2003.

\bibitem{Garcia:Salvador:2010}
S.~Garc{\'\i}a, A.~Fern{\'a}ndez, J.~Luengo, F.~Herrera, Advanced nonparametric
  tests for multiple comparisons in the design of experiments in computational
  intelligence and data mining: Experimental analysis of power, Information
  Sciences 180~(10) (2010) 2044--2064.

\end{thebibliography}

\newcommand{\noop}[1]{}

\end{document}